\documentclass{article}

\PassOptionsToPackage{numbers, square, sort}{natbib}
\usepackage[preprint]{neurips_2020}
\usepackage[utf8]{inputenc} 
\usepackage[T1]{fontenc}    
\usepackage{hyperref}       
\usepackage{url}            
\usepackage{booktabs}       
\usepackage{amsfonts}       
\usepackage{nicefrac}       
\usepackage{microtype}      
\usepackage{multirow}       

\usepackage{graphicx} 

\usepackage{xcolor}

\newif\ifcomments
\commentsfalse 

\ifcomments
\newcommand{\aga}[1]{{\color{red}(#1)}} 
\newcommand{\scom}[1]{{\color{blue}(#1)}}
\newcommand{\jp}[1]{{\color{purple}(jp: #1)}}
\newcommand{\todo}[1]{{\color{gray}TODO: #1}} 
\else
\newcommand{\aga}[1]{}
\newcommand{\scom}[1]{}
\newcommand{\jp}[1]{}
\newcommand{\todo}[1]{}
\fi

\usepackage{amsmath}
\usepackage{mathmacros}

\renewcommand{\th}{\sm{\theta}} 
\newcommand{\ntk}{\hat{\Theta}} 
\newcommand{\ntkz}{\m{Id}_{\z}\otimes \ntk} 

\newcommand{\soft}{\sigma} 
\newcommand{\lr}{\eta} 
\newcommand{\tlr}{\tilde{\lr}} 
\newcommand{\dsoft}{\sm{\soft}_{z}} 
\newcommand{\X}{\mathcal{X}} 
\newcommand{\x}{\m{x}} 
\newcommand{\Y}{\mathcal{Y}} 
\newcommand{\Lf}{\mathcal{L}} 
\newcommand{\z}{\m{z}} 
\newcommand{\Z}{\m{Z}} 
\newcommand{\Zparam}{\lVert\Z^{0}\rVert_{F}} 
\newcommand{\g}{\gamma} 
\renewcommand{\c}{c} 
\newcommand{\K}{K} 
\newcommand{\M}{M} 
\renewcommand{\P}{P} 
\newcommand{\N}{N} 
\newcommand{\zs}{\z^*} 
\newcommand{\dz}{\delta\z} 

\newcommand{\T}{T}
\newcommand{\tz}{\tau_{z}}
\newcommand{\tnl}{\tau_{nl}}

\renewcommand{\v}{\m{v}}
\newcommand{\gd}{\m{g}}
\newcommand{\lam}{\lambda}
\newcommand{\dth}{\Delta_{\th}}
\newcommand{\dlf}{\Delta_{\Lf}}
\newcommand{\nub}{\sm{\nu}}

\newcommand{\h}{\m{h}}
\newcommand{\W}{\m{W}}

\renewcommand{\H}{\m{H}}
\newcommand{\sign}{{\rm sign}}

\title{\aga{Comments on}Temperature check: theory and practice for training models with softmax-cross-entropy losses}

\author{%
Atish Agarwala \\
 Google Research\\
  \texttt{thetish@google.com} \\
  \And
Jeffrey Pennington \\
 Google Research\\
  \texttt{jpennin@google.com} \\
  \And
Yann Dauphin \\
 Google Research\\
  \texttt{ynd@google.com} \\
    \And
Sam Schoenholz \\
 Google Research\\
  \texttt{schsam@google.com} \\
}

\begin{document}

\maketitle

\begin{abstract}
The softmax function combined with a cross-entropy loss is a principled approach to modeling probability distributions that has become ubiquitous in deep learning. The softmax function is defined by a lone hyperparameter, the temperature, that is commonly set to one or regarded as a way to tune model confidence after training; however, less is known about how the temperature impacts training dynamics or generalization performance. In this work we develop a theory of early learning for models trained with softmax-cross-entropy loss and show that the learning dynamics depend crucially on the inverse-temperature $\beta$ as well as the magnitude of the logits at initialization, $||\beta\z||_{2}$. We follow up these analytic results with a large-scale empirical study of a variety of model architectures trained on CIFAR10, ImageNet, and IMDB sentiment analysis. We find that generalization performance depends strongly on the temperature, but only weakly on the initial logit magnitude. We provide evidence that the dependence of generalization on $\beta$ is not due to changes in model confidence, but is a dynamical phenomenon. It follows that the addition of $\beta$ as a tunable hyperparameter is key to maximizing model performance. Although we find the optimal $\beta$ to be sensitive to the architecture, our results suggest that tuning $\beta$ over the range $10^{-2}$ to $10^1$  improves performance over all architectures studied. We find that smaller $\beta$ may lead to better peak performance at the cost of learning stability.
\end{abstract}

\section{Introduction}

Deep learning has led to 
breakthroughs across a slew of classification 
tasks~\citep{lecun_backpropagation_1989,krizhevsky_imagenet_2012, zagoruyko_wide_2017}.
Crucial components of this success have been the use of the softmax 
function to model predicted class-probabilities combined with
the cross-entropy loss function as a measure of distance between the predicted distribution 
and the label \citep{kline_revisiting_2005, golik_crossentropy_2013}.
Significant work has gone into improving the generalization
performance of softmax-cross-entropy learning. A particularly successful 
approach has been to improve overfitting by reducing model confidence; this 
has been done by regularizing outputs using confidence 
regularization~\citep{pereyra_regularizing_2017} 
or by augmenting data using label 
smoothing~\citep{muller_when_2019,szegedy_rethinking_2016}. Another way to manipulate model 
confidence is to tune the temperature of the softmax function, which is otherwise commonly set to one. 
Adjusting the softmax temperature during training has been shown to be important in metric 
learning~\citep{wu_improving_2018,zhai_classification_2019} and when performing 
distillation~\citep{hinton_distilling_2015}; as well as for post-training
calibration of prediction probabilities 
\citep{platt_probabilistic_2000,guo_calibration_2017}.

The interplay between temperature, learning, and generalization
is complex and not well-understood in the general case.
Although significant recent theoretical progress has been made understanding 
generalization and learning in wide neural networks approximated as linear models, 
analysis of linearized learning dynamics
has largely focused on the case of squared error losses~
\citep{jacot_neural_2018, du_gradient_2019, lee_wide_2019, novak_bayesian_2019,xiao_disentangling_2019}.
Infinitely-wide networks trained with softmax-cross-entropy loss have been shown to
converge to max-margin classifiers in a particular function space norm \citep{chizat_implicit_2020},
but timescales of convergence are not known.
Additionally, many well-performing models operate best away from the
linearized regime
\citep{aitchison_why_2019,novak_bayesian_2019}.
This means that understanding the deviations of models from their linearization
around initialization is important for understanding generalization
\citep{lee_wide_2019,chizat_lazy_2019}.

In this paper, we investigate the training of neural networks with softmax-cross-entropy losses. 
In general this problem is analytically intractable; to make progress we pursue a strategy
that combines analytic insights at short times with a comprehensive set of experiments that capture
the entirety of training. At short times, models can be understood in terms of a linearization
about their initial parameters along with nonlinear corrections. In the linear regime we find that
networks trained with different inverse-temperatures, $\beta = 1 / T$,  behave identically provided
the learning rate is scaled as $\tlr = \lr \beta ^2$. Here, networks begin to learn over a timescale 
$\tz\sim \lVert\Z^{0}\rVert_{2} / \tlr$ where $\Z^{0}$ are the initial
logits of the network after being multiplied by $\beta$. 
This implies that we expect learning to begin faster for networks with smaller logits.
The learning dynamics begin to become nonlinear over another, independent, timescale $\tnl\sim \beta/\tlr$, 
suggesting more nonlinear learning for small $\beta$.
From previous results we expect that neural networks will perform best in this regime
where they quickly exit the linear regime \citep{chizat_lazy_2019,lee_finite_2020,lewkowycz_large_2020}.

We combine these analytic results with extensive experiments on competitive neural networks across a 
range of architectures and domains including: Wide Residual networks~\citep{zagoruyko_wide_2017} on 
CIFAR10~\citep{krizhevsky_learning_2009}, ResNet-50~\citep{he_deep_2016} 
on ImageNet~\citep{deng_imagenet_2009}, and GRUs~\citep{chung_empirical_2014} on the IMDB sentiment analysis 
task~\citep{maas_learning_2011}. In the case of residual networks, 
we consider architectures with and without batch normalization, which can appreciably change the learning 
dynamics~\citep{ioffe_batch_2015}. 
For all models studied, we find that generalization performance is poor at $\lVert\Z^{0} \rVert_{2}\gg 1$ but otherwise
largely independent of $\lVert\Z^{0} \rVert_{2}$.
Moreover, 
learning 
becomes slower and less stable at very
small $\beta$; indeed, the optimal learning rate scales like $\lr^*\sim 1/\beta$ 
and the resulting early learning timescale can be written as $\tz^*\sim \lVert\Z^{0} \rVert_{2} / \beta$. For all models 
studied,
we observe strong performance for $\beta\in[10^{-2},10^1]$ although the specific optimal $\beta$ is architecture 
dependent. Emphatically, the optimal $\beta$ is often far from 1. For models without batch normalization, smaller 
$\beta$
can give stronger results on some training runs, with others failing to train due to instability.
Overall, these results suggest 
that model performance can often be improved
by tuning $\beta$ over the range of $[10^{-2},10^1]$. 

\section{Theory}

We begin with a precise description of the problem setting before discussing a theory of learning at short times.

\label{sec:theory}
\subsection{Basic model and notation}

We consider a classification task with $\K$ classes. For an $\N$ dimensional input $\x$, let $\z(\x,\th)$ be the pre-softmax output
of a classification model parameterized by
$\th\in\mathbb{R}^\P$, such that the classifier predicts the class $i$ corresponding to the largest output
value $\z_{i}$.
We will mainly consider $\th$ trained by SGD on a training set $(\X,\Y)$ of $\M$ input-label
pairs. We focus on models trained with cross-entropy loss with a
non-trivial \emph{inverse temperature} $\beta$.
The softmax-cross-entropy loss can be written as
\begin{equation}
\Lf(\th,\X,\Y) = \sum_{i=1}^{\K}\Y_{i}\cdot\ln(\soft(\beta \z_{i}(\X,\th))) = \sum_{i=1}^K\Y_{i}\cdot\ln(\soft(\Z_{i}(\X,\th)))
\end{equation}
where we define the \emph{rescaled logits} $\Z=\beta\z$
and 
$\soft(\Z)_{i} = e^{\Z_{i}}/\sum_{j}e^{\Z_{j}}$ is the softmax function. Here $\Z(\X,\th)$ is
the $\M \times \K$ dimensional matrix of rescaled logits on the training set.

As we will see later, the statistics of individual $\soft(\Z)_{i}$ will have a strong influence on the
learning dynamics. While the statistics of $\soft(\Z)_i$ are intractable for intermediate
magnitudes, $\lVert\Z\rVert_{2}$, they can be understood in the limits of large and small $\lVert\Z\rVert_{2}$.
For a fixed model $\z(\x,\th)$, $\beta$ controls the certainty of the predicted
probabilities. Values of $\beta$ such that $\beta\ll1/\lVert\z\rVert_{2}$ will 
give small values of $\lVert\Z\rVert_{2} \ll 1$, and the
outputs of the softmax 
will be close to $1/\K$ independent of $i$ (the maximum-entropy distribution on $K$ classes). Larger 
values of $\beta$ such that $\beta\gg 1/\lVert\z\rVert_{2}$ will lead to large values of 
$\lVert\Z\rVert_{2}\gg 1$; the resulting distribution has probability
close to $1$ on one label, and (exponentially) close to $0$ on the others.

The continuous time learning dynamics
(exact in the limit of small learning rate) are given by:
\begin{equation}
\dot{\th}  = \lr\beta \sum_{i=1}^{\K}\left(\frac{\partial \z_{i}(\X,\th(t))}{\partial\th}\right)^{\tpose}(\Y_{i}-\soft(\Z_{i}(\X,\th(t)))
\end{equation}
for learning rate $\lr$. We will drop the explicit dependence of $\Z_{i}$ on $\th$
from here onward, and we will denote time dependence as $\Z_{i}(\X,t)$ explicitly where needed.

In function space, the dynamics of the model outputs
on an input $\x$ are given by
\begin{equation}
\frac{d\z_{i}(\x)}{dt} = \lr\beta \sum_{j=1}^{\K}(\ntk_{\th})_{ij}(\x,\X)(\Y_j-\soft(\Z_j(\X)))
\label{eq:z_ntk}
\end{equation}
where we define the $\M\times\M\times\K\times\K$ dimensional tensor $\ntk_{\th}$,
the empirical NTK, as
\begin{equation}
(\ntk_{\th})_{ij}(\x,\X')\equiv \frac{\partial\z_{i}(\x)}{\partial\th}\left(\frac{\partial\z_{j}(\X)}{\partial\th}\right)^{\tpose}
\end{equation}
for class indices $i$ and $j$, which is block-diagonal in the infinite-width limit.

From Equation \ref{eq:z_ntk}, we see that the early-time dynamics in function
space depend on $\beta$, the initial softmax input $\Z(\X,0)$ on the training set, and the initial
$\ntk_{\th}$. Changing these observables across a model family will lead to
different learning trajectories early in learning. Since significant work has already studied the effects of the NTK, here we focus on
the effects of changing $\beta$ and $\lVert\Z^{0}\rVert_{F}\equiv \lVert\Z(\X,0)\rVert_{F}$
(the norm of the $\M\times \K$ dimensional matrix of training logits), independent of
$\ntk_{\th}$.

\subsection{Linearized dynamics}

\label{sec:lin-time}

For small changes in $\th$, the tangent kernel is approximately constant throughout
learning \citep{jacot_neural_2018}, and 
we drop the explicit $\th$
dependence in this subsection.
The linearized dynamics of $\z(\x,t)$ only depend on the initial
value of $\ntk$ and the $\beta$-scaled logit values
$\Z(\X,t)$. This suggests that there is a universal timescale
across $\beta$ and $\lr$ which can be used to compare linearized trajectories
with different parameter values. Indeed, if we define an \emph{effective
learning rate} $\tlr \equiv \lr\beta^2$, we have
\begin{equation}
\frac{d\Z_i(\x)}{dt} = \tlr\sum_{j=1}^{\K} (\ntk)_{ij}(\x,\X)(\Y_{j}-\soft(\Z_{j}(\X)))
\label{eq:rescaled_linear}
\end{equation}
which removes explicit $\beta$ dependence of the dynamics.
We note that a similar rescaling exists for the continuous time
versions of other optimizers like momentum 
(Appendix \ref{app:mom_rescaling}).

\begin{figure}[h]
\centering
\begin{tabular}{cc}
\includegraphics[height=0.25\linewidth]{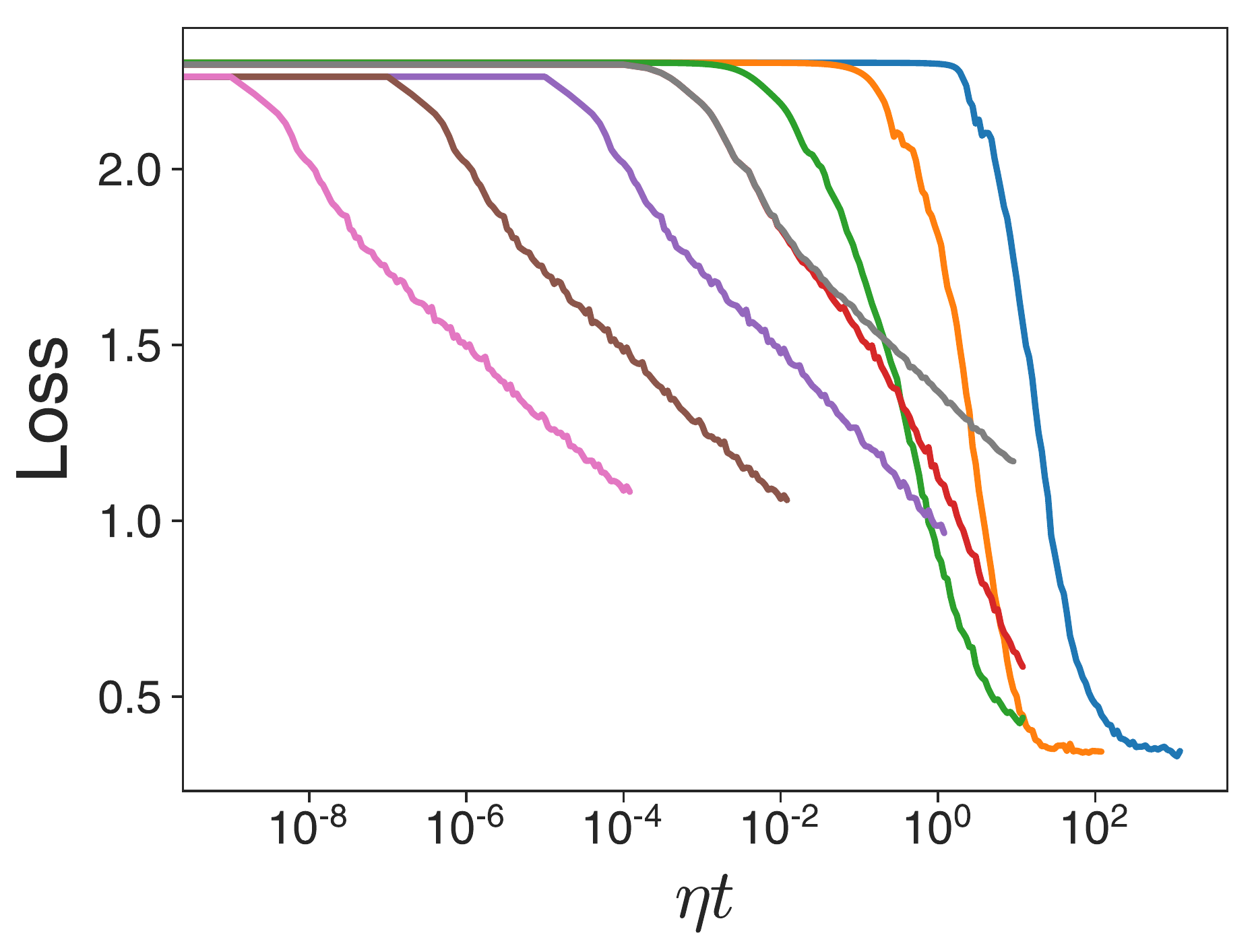} & \includegraphics[height=0.25\linewidth]{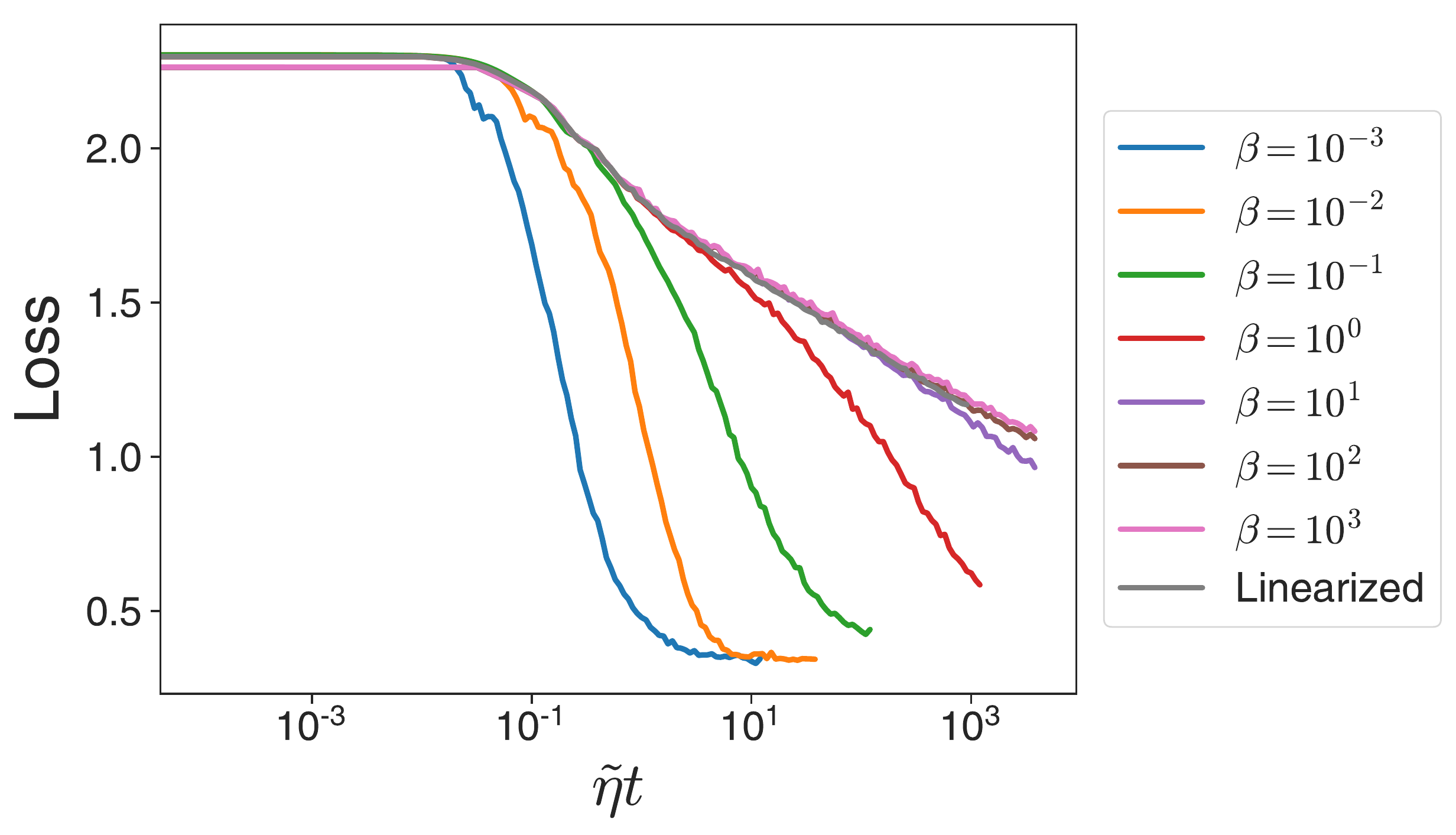}
\end{tabular}
\caption{For fixed initial training set logits $\Z^{0}$,
plotting learning curves against $\tlr t = \beta^2\lr t$
causes the learning curves to collapse to the learning curve of the linearized model at early
times (right), in contrast to un-scaled curves (left). Models with large $\beta$
follow linearized dynamics the longest.}
\label{fig:loss_collapse}
\end{figure}

The effective learning rate $\tlr$ is useful for understanding the nonlinear dynamics,
as plotting learning curves versus $\tlr t$ causes early-time collapse for fixed
$\Z^{0}$ across $\beta$ and $\lr$ (Figure \ref{fig:loss_collapse}).
We see that there is a strong, monotonic, dependence of the time at which the nonlinear model 
begins to deviate from its linearization on $\beta$. We will return to and explain this phenomenon in Section 
\ref{sec:non-lin}.

Unless otherwise noted, we will analyze all timescales in units of $\tlr$ instead of
$\lr$, as it will allow for the appropriate early-time comparisons between
models with different $\beta$.

\subsection{Early learning timescale}
\label{sec:early_learning}

We now define and compute the early learning timescale, $\tz$, that measures the time it takes for the logits 
to change significantly from their initial value. Specifically, we define
$\tz$ such that for $t \ll \tz$ we expect $\lVert\Z(\x,t) - \Z(\x, 0)\rVert_{F}\ll \lVert\Z(\x, 0)\rVert_{F}$ and for 
$t\gg\tz$, $\lVert\Z(\x, t) - \Z(\x, 0)\rVert_{F} \sim \lVert \Z(\x, 0) \rVert_{F}$ (or larger).
This is synonymous with the timescale
over which the model begins to learn. As we will show below,
$\tz\propto \lVert\Z^{0}\rVert_{F}/\tlr$. Therefore in units
of $\tlr$, $\tz$ only depends on $\lVert\Z^{0}\rVert_{F}$ and not $\beta$.

To see this, note that at very short times it follows from Equation~\ref{eq:rescaled_linear} that 
\begin{equation}
\Z_{i}(\x, t) - \Z_{i}(\x, 0) \approx \tlr\sum_{j=1}^{\K}(\ntk)_{ij}(\x,\X)(\Y_j-\soft(\Z_j(\X)))t + \mathcal O(t^2)
\label{eq:first_order}
\end{equation}
It follows that we 
can define a timescale over which the logits (on the training set)
change appreciably from their initial value as
\begin{equation}
\tz \equiv \frac{1}{\tlr}\frac{\lVert\Z^{0}\rVert_{F}}{\lVert \ntk(\X,\X)(\Y-\soft(\Z^{0})) \rVert_{F}}.
\label{eq:tau_z_eq}
\end{equation}
where the norms are once again taken across all classes as well as training points. This definition has the desired properties for $t\ll\tz$ and $t\gg\tz$.

In units of $\tlr$, $\tz$ depends only on $\Zparam$, in two ways.
The first is a linear scaling in $\lVert\Z^{0}\rVert_{F}$; the second comes from the
contribution from the gradient
$\lVert \ntk(\X,\X)(\Y-\soft(\Z(\X,0)))\rVert_{F}$. As previously discussed, since
$\sigma(\Z^0)$ saturates 
at small and large values of $\lVert\Z^{0}\rVert_{F}$, it follows that the gradient term will also 
saturate for large and small $\Zparam$, and the ratio of saturating values is some $O(1)$
constant independent of 
$\Zparam$ and $\beta$.

\begin{figure}[h]
\centering
\begin{tabular}{ccc}
\includegraphics[width=0.31\linewidth]{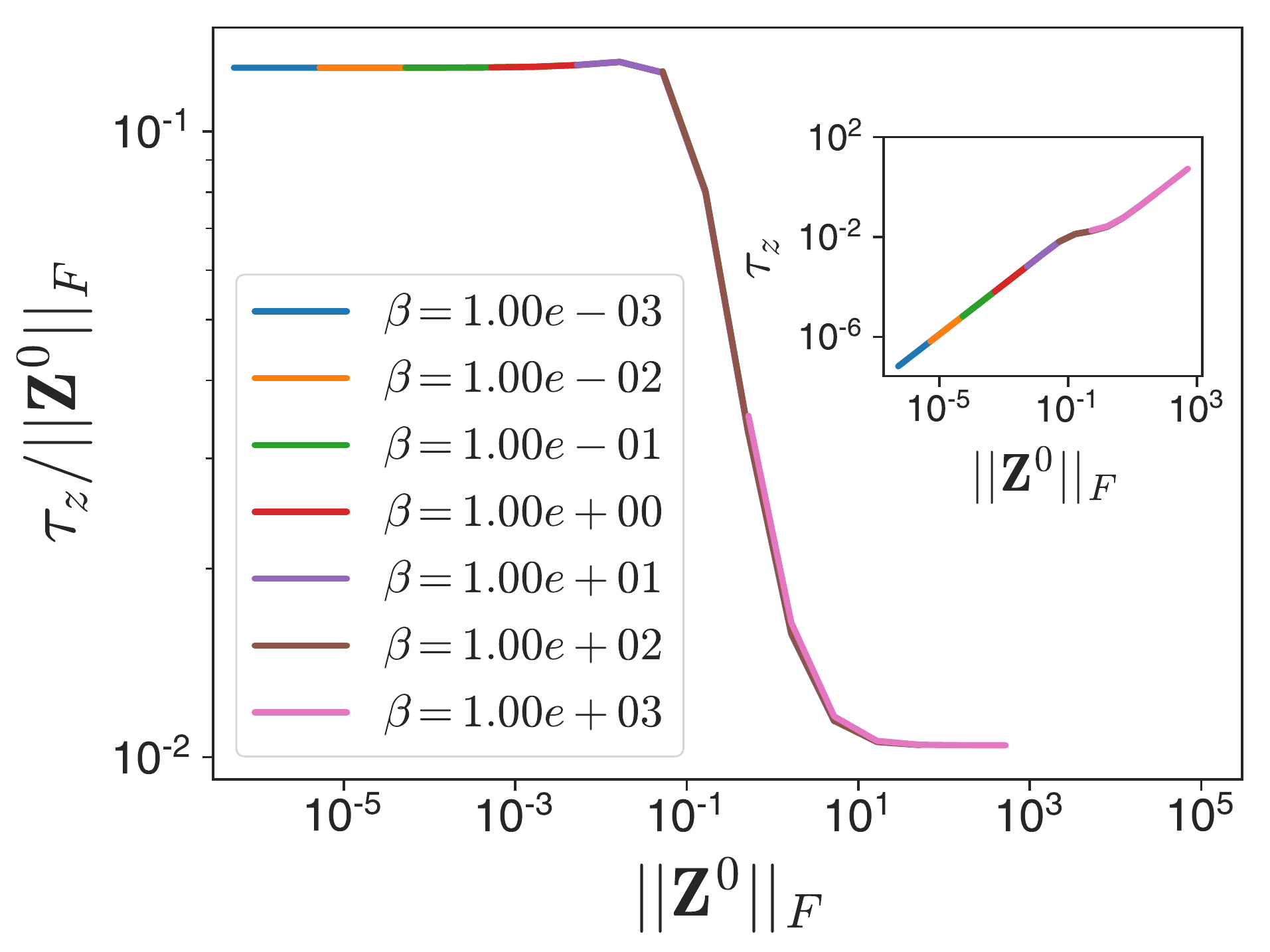} &  
\includegraphics[width=0.31\linewidth]{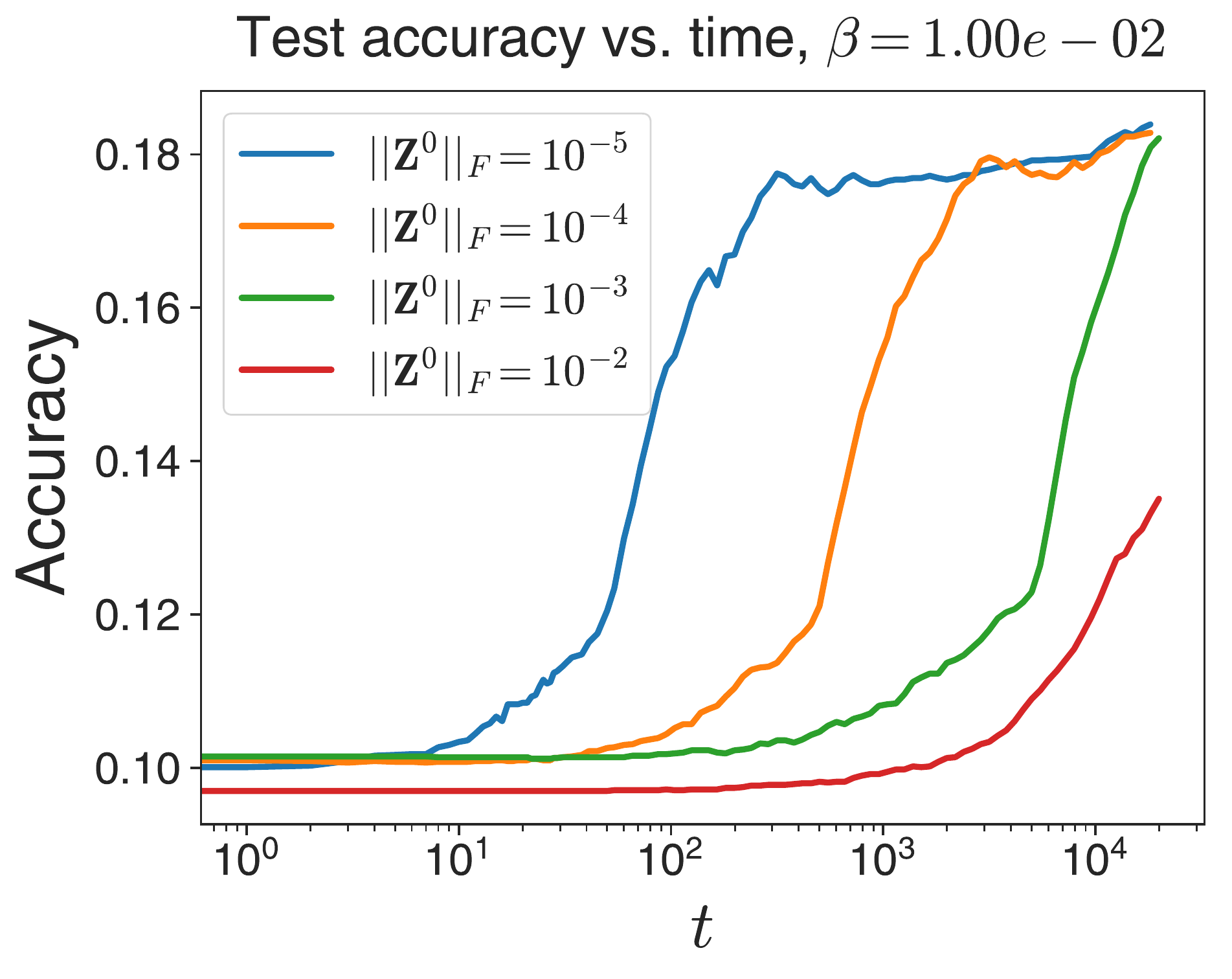} &
\includegraphics[width=0.31\linewidth]{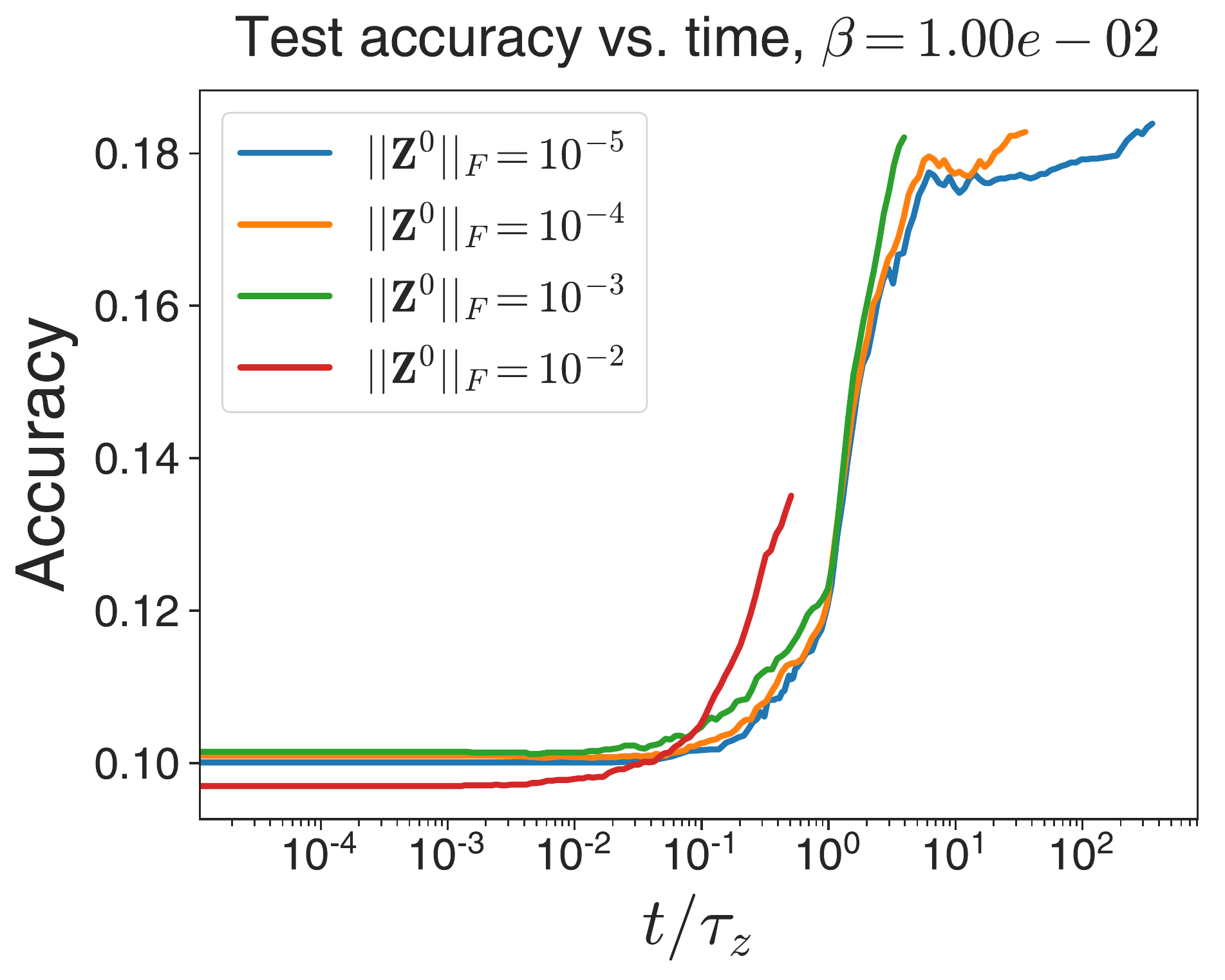}\\
\end{tabular}
\caption{The timescale $\tz$
depends only on $\Zparam$ in units of $\tlr = \beta^2\lr$ (left, inset). $\tz$ depends
linearly on
$\lVert\Z^{0}\rVert_{F}$,
up to an $O(1)$ coefficient which saturates at large and small $\lVert\Z^{0}\rVert_{F}$ (left, main).
Accuracy increases more quickly for
small initial $\lVert\Z^{0}\rVert_{F}$, though late time dynamics are similar (center). Rescaling time to $t/\tz$ causes early
accuracy curves to collapse (right).}
\label{fig:tau_1_plots}
\end{figure}

The quantitative and conceptual nature of $\tz$ can both be confirmed numerically. When
plotted over a wide range of $\lVert\Z^{0}\rVert_{F}$ and $\beta$, the ratio $\tz/\lVert\Z^{0}\rVert_{F}$
(in rescaled time units)
undergoes a saturating, $O(1)$ variation from small to large $\lVert\Z^{0}\rVert_{F}$ (Figure 
\ref{fig:tau_1_plots}, left). The quantitative dependence of the transition
on the NTK is confirmed in Appendix \ref{app:Z_dep}. Additionally, for fixed
$\beta$ and varying $\lVert\Z^{0}\rVert_{F}$, rescaling time by $1/\tz$ causes accuracy curves
to collapse at early times (Figure \ref{fig:tau_1_plots}, middle), even if they
are very different at early times without the rescaling (right). We note here
that the late time accuracy curves seem similar across $\lVert\Z^{0}\rVert_{F}$ without rescaling,
a point which we will return to in Section \ref{sec:late_time}.

\subsection{Nonlinear timescale}

\label{sec:non-lin}

While linearized dynamics are useful to understand some features of learning,
the best performing networks often reside in the nonlinear 
regime~\citep{novak_bayesian_2019}. Here we define the nonlinear timescale,
$\tnl$, corresponding to the time over which the network deviates appreciably from the linearized equations. We will show that
$\tnl\propto \beta/\tlr$.
Therefore, in terms of $\beta$ and $\lVert\Z^{0}\rVert_{F}$, networks with small $\beta$ will
access the nonlinear regime early in learning, while networks with large $\beta$
will be effectively linearized throughout training. We note that a similar
point was raised in \citet{chizat_lazy_2019}, primarily in the context of MSE loss.

We define $\tnl$ to be the timescale over which the change in $\ntk_{\th}$
(which contributes to
the second order term in Equation \ref{eq:first_order}) can no longer be neglected.
Examining the second time derivative of $\Z$, we have
\begin{equation}
\frac{d^2\Z_{i}}{dt^2} = \tlr\sum_{j=1}^{\K} \left(
\underbrace{-(\ntk_{\th})_{ij}(\x,\X)\frac{d}{dt}\soft(\Z(\X))}_{\text{linearized~dynamics}}
+\underbrace{\frac{d}{dt}\left[(\ntk_{\th})_{ij}(\x,\X)\right](\Y_{j}-\soft(\Z_{j}(\X))}_{\text{nonlinearized~dynamics }\equiv\, \ddot{\Z}_{nl}}\right)
\end{equation}
The first term is the second derivative under a fixed kernel, while the second term is due to the change in the kernel
(neglected in the linearized limit). A direct calculation shows that the second term, which we denote $\ddot{\Z}_{nl}$,
can be written as
\begin{equation}
(\ddot{\Z}_{nl})_{i} = \beta^{-1}\tlr^{2}\sum_{j=1}^{\K}\sum_{k=1}^{\K} (\Y_{k}(\X)-\soft(\Z_{k}(\X))^{\tpose}\left(\frac{\partial \z_{k}(\X)}{\partial\th}\cdot\frac{\partial}{\partial\th}
[\ntk_{\th}]_{ij}\right)(\Y_{j}(\X)-\soft(\Z_{j}(\X)))
\label{eq:Z_nl}
\end{equation}
This gives us a \emph{nonlinear timescale} $\tnl$ defined, at initialization, by $\tnl \equiv \lVert\dot{\Z}(\X,0)\rVert_{F}/\lVert\ddot{\Z}_{nl}(\X,0)\rVert_{F}$.
We can interpret $\tnl$ as the time it takes for changes in the kernel to
contribute to learning.

Though computing $\lVert\ddot{\Z}_{nl}(\X,0)\rVert_{F}$ in exactly is analytically intractable,
its basic scaling in terms of $\beta$ and $\Zparam$ (and therefore, that of
$\tnl$) is computable. We first note the explicit $\beta^{-1}\tlr^{2}$ dependence. 
The remaining terms are independent of $\beta$ and vary by at most $\mathcal O(1)$ with $\Zparam$; indeed as described above,
$\lVert\Y(\X)-\soft(\Z(\X,0))\rVert_{F}$ saturates for large and small $\lVert\Z^{0}\rVert_{F}$. Morevoer, the derivative,
$\frac{\partial\z(\X,0)}{\partial \th}$, is the square root of the NTK and, at initialization, it
is independent of $\lVert\Z^{0}\rVert_{F}$. Together
with our analysis of $\tz$ we have that, up to some $O(1)$
dependence on $\lVert\Z^{0}\rVert_{F}$, $\tnl\propto \beta/\tlr$. Therefore, the degree of nonlinearity early in learning
is controlled via $\beta$ alone.

\begin{figure}[h]
\centering
\begin{tabular}{cc}
\includegraphics[width=0.4\linewidth]{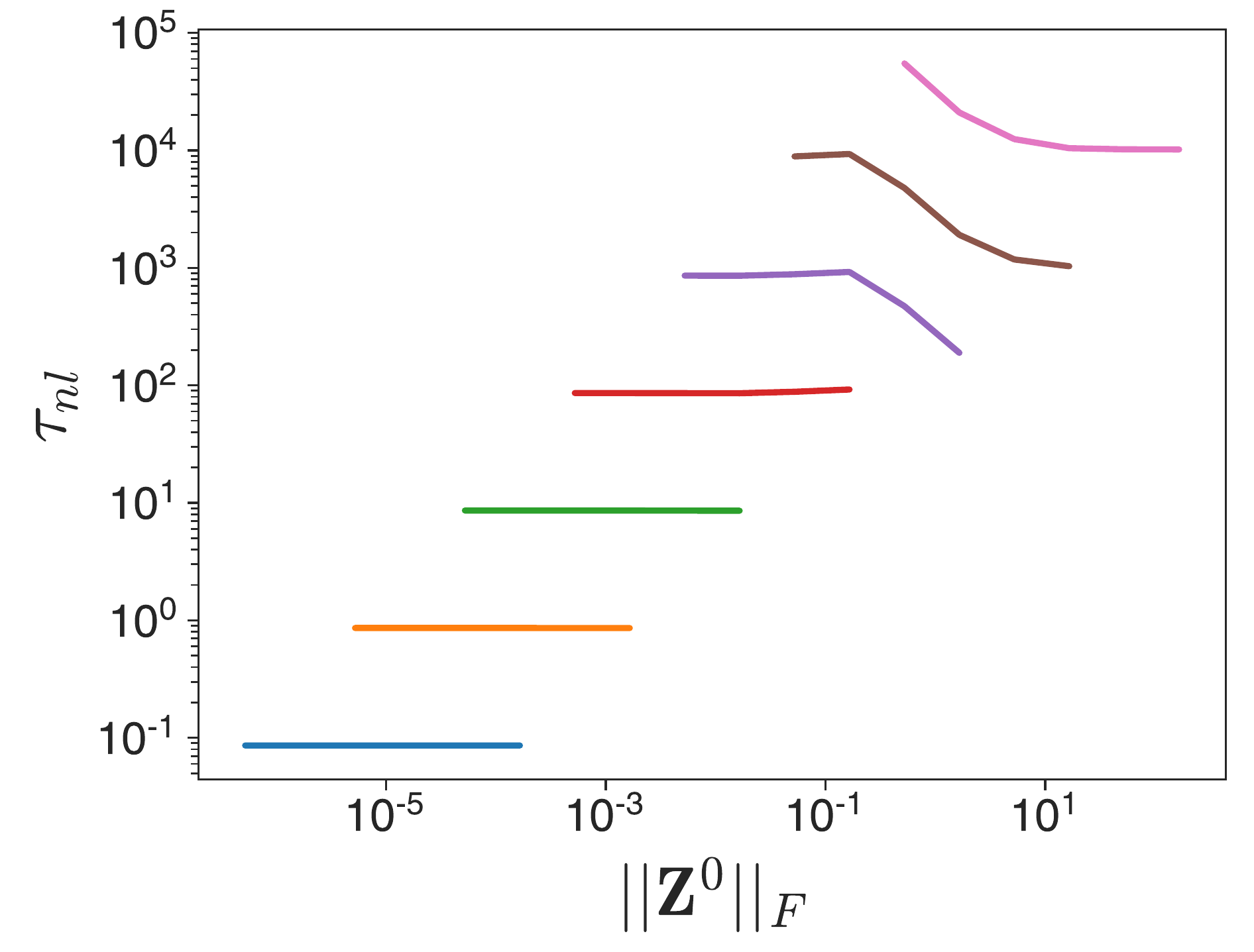} & \includegraphics[width=0.4\linewidth]{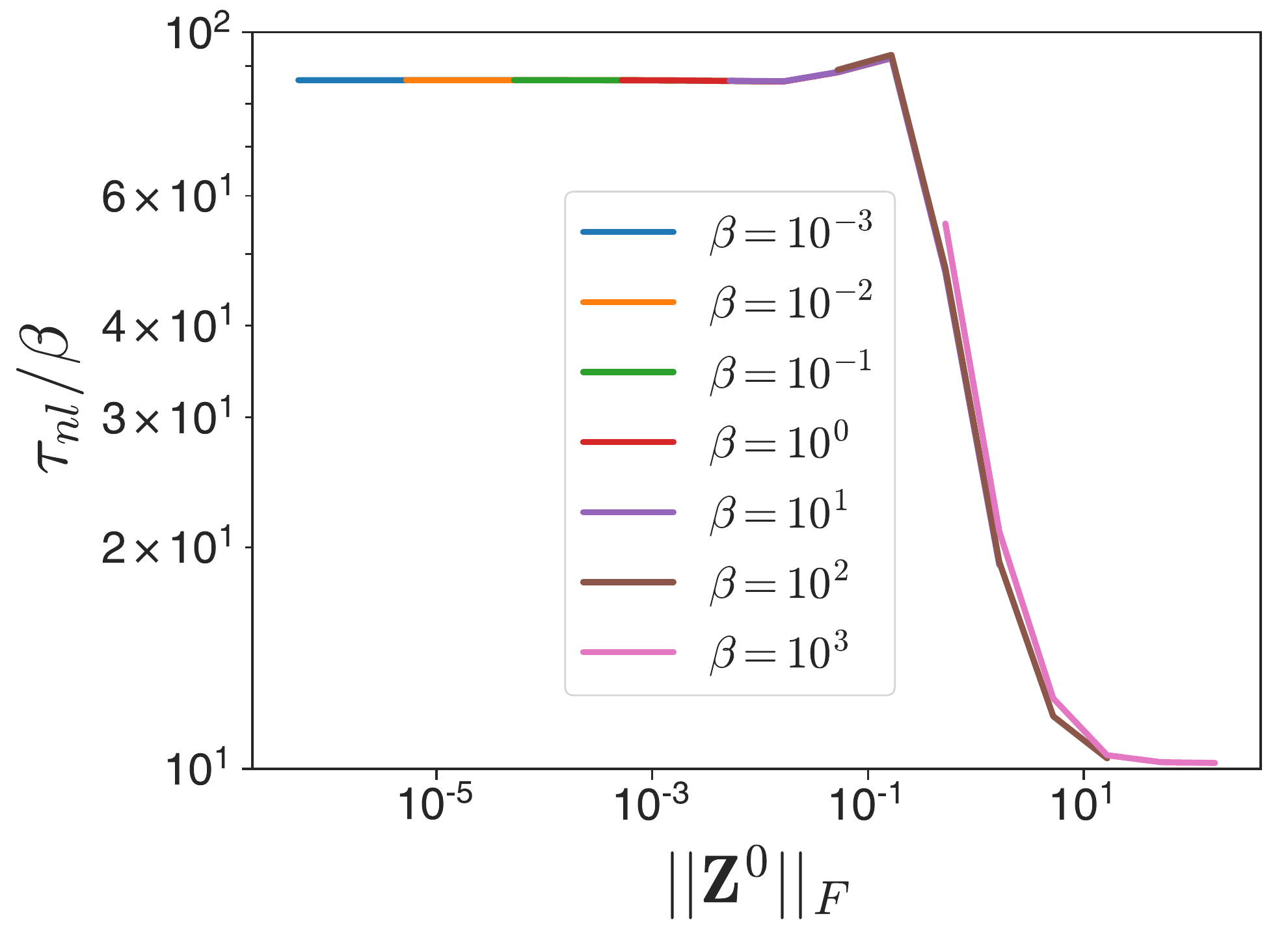}
\end{tabular}
\caption{The time to deviation from linearized dynamics, $\tau_{nl}$,
has large deviation over $\beta$ and $\lVert\Z^{0}\rVert_{F}$ (left),
which can be largely explained by linear dependence on $\beta$ (right),
in units of $\tlr = \beta^2\lr$. There is an $O(1)$
dependence on $\lVert\Z^{0}\rVert_{F}$
which is consistent across varying $\beta$ for fixed $\lVert\Z^{0}\rVert_{F}$.}
\label{fig:tau_nl}
\end{figure}

Once again we can confirm the quantitative and conceptual understanding of $\tnl$
numerically. Qualitatively, we see that for fixed $\lVert\Z^{0}\rVert_{F}$, models with smaller
$\beta$ deviate sooner from the linearized dynamics when learning curves are
plotted against $\tlr t$ (Figure \ref{fig:loss_collapse}).
Quantitatively, we see that $\tnl/\beta$ (in units of $\tlr$) has an $O(1)$
dependence on $\lVert\Z^{0}\rVert_{F}$ only (Figure \ref{fig:tau_nl}).

\section{Experimental Results}


\subsection{Optimal Learning Rate}

\begin{figure}[t]
\centering
\includegraphics[width=0.4\linewidth]{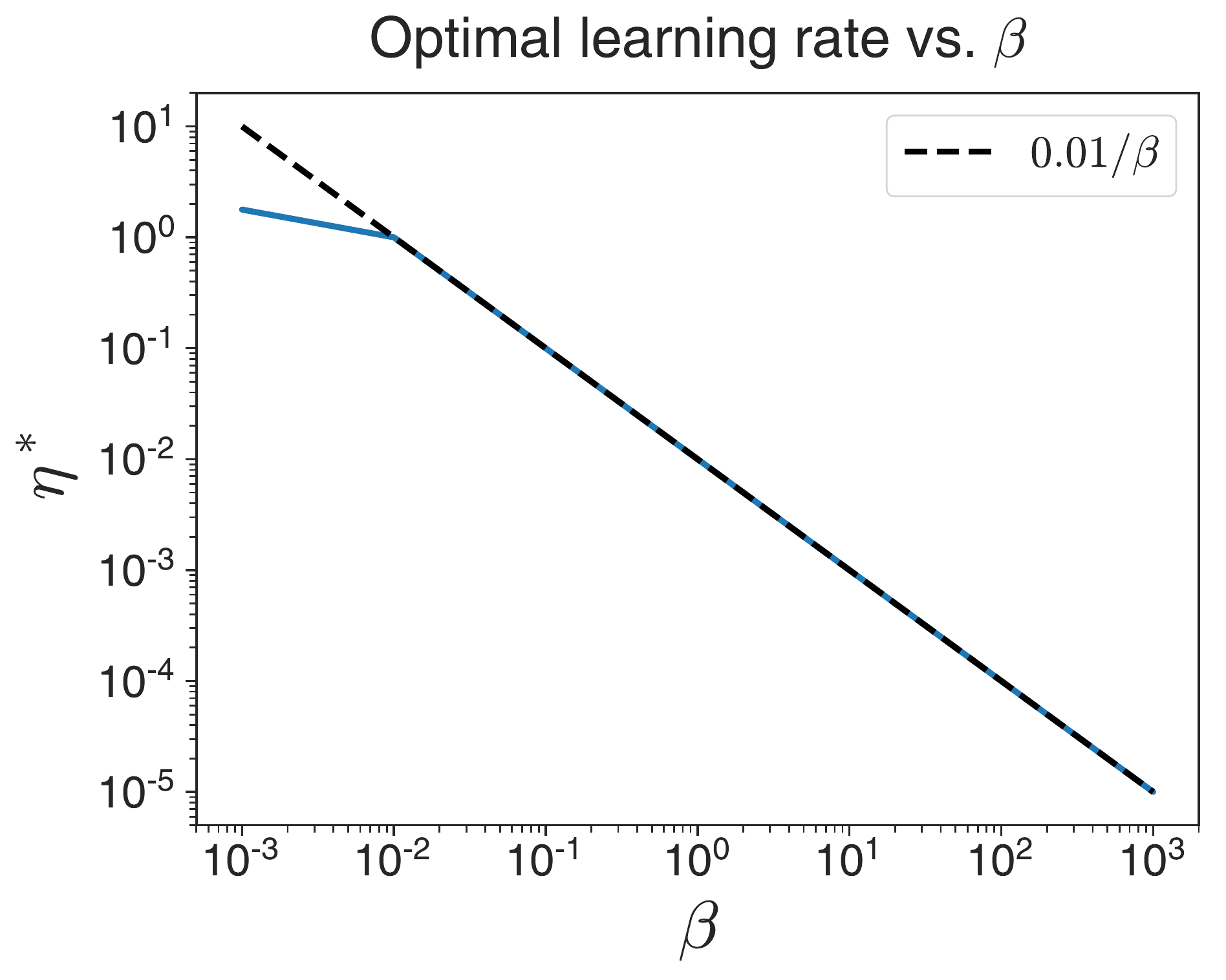}
\caption{Optimal learning rate $\lr^*$ for WRN on CIFAR10 scales as $1/\beta$.}
\label{fig:learning_rate}
\end{figure}

We begin our empirical investigation by training wide resnets~\citep{szegedy_rethinking_2016} without batch 
normalization on CIFAR10. In order to understand the effects of the different timescales on learning,
we control $\beta$ and $\Zparam$ independently by using a correlated initialization strategy outlined in 
Appendix~\ref{app:cor_init}.

Before considering model performance, it is first useful to understand the scaling of the optimal learning rate with $\beta$. To do this, we initialize networks with different $\beta$ and conduct learning rate 
sweeps 
for each 
$\beta$. The optimal learning rate $\lr^*$ has a clear $1/\beta$ dependence (Figure~\ref{fig:learning_rate}).
Plugging this optimal learning rate into the two timescales identified above gives
$\tz^* \sim \Zparam / \beta$ and $\tnl^*\sim \mathcal O(1)$. Note that these timescales are now in units of
SGD steps. This suggests that the maximum learning rate 
is set so that nonlinear effects become important at the fastest possible rate without leading to instability. We notice 
that $\tau_z$ will be large at small $\beta$ and small at large $\beta$. Thus, at small $\beta$ we expect learning to 
take place slowly and nonlinear effects to become important by the time the function has changed appreciably. At large 
$\beta$, by contrast, our results suggest that the network will have learned a significant amount before the dynamics 
become appreciably nonlinear.

\subsection{Phase plane}

\label{sec:late_time}

In the preceding discussion two quantities emerged that control the behavior of early-time dynamics: the 
inverse-temperature, $\beta$, and the rescaled logits $\Zparam$. In attempting to understand the behavior of real neural 
networks trained using softmax-cross-entropy loss, it therefore makes sense to try to reason about this behavior by 
considering neural networks that span the $\beta-\lVert\Z^{0}\rVert_{F}$
\emph{phase plane}, the space
of allowable pairs $(\beta,\Zparam)$. By construction, the phase plane is characterized by the timescales involved in 
early learning. To summarize, $\tz\sim\lVert\Z^{0}\rVert_{F}/\tlr$ sets the
timescale for early learning, with larger values of
$\lVert\Z^{0}\rVert_{F}$ leading to longer time before significant accuracy
gains are made (Section \ref{sec:early_learning}).
Meanwhile, $\tnl\sim\beta/\tlr$ controls the timescale for learning dynamics to leave
the linearized regime - with small $\beta$ leading to immediate departures
from linearity, while models with large $\beta$ may stay linearized throughout
their learning trajectories (Section \ref{sec:non-lin}).

\begin{figure}[h]
\centering

\begin{tabular}{cc}

\multirow{2}{*}[18mm]{\includegraphics[width=0.55\linewidth]{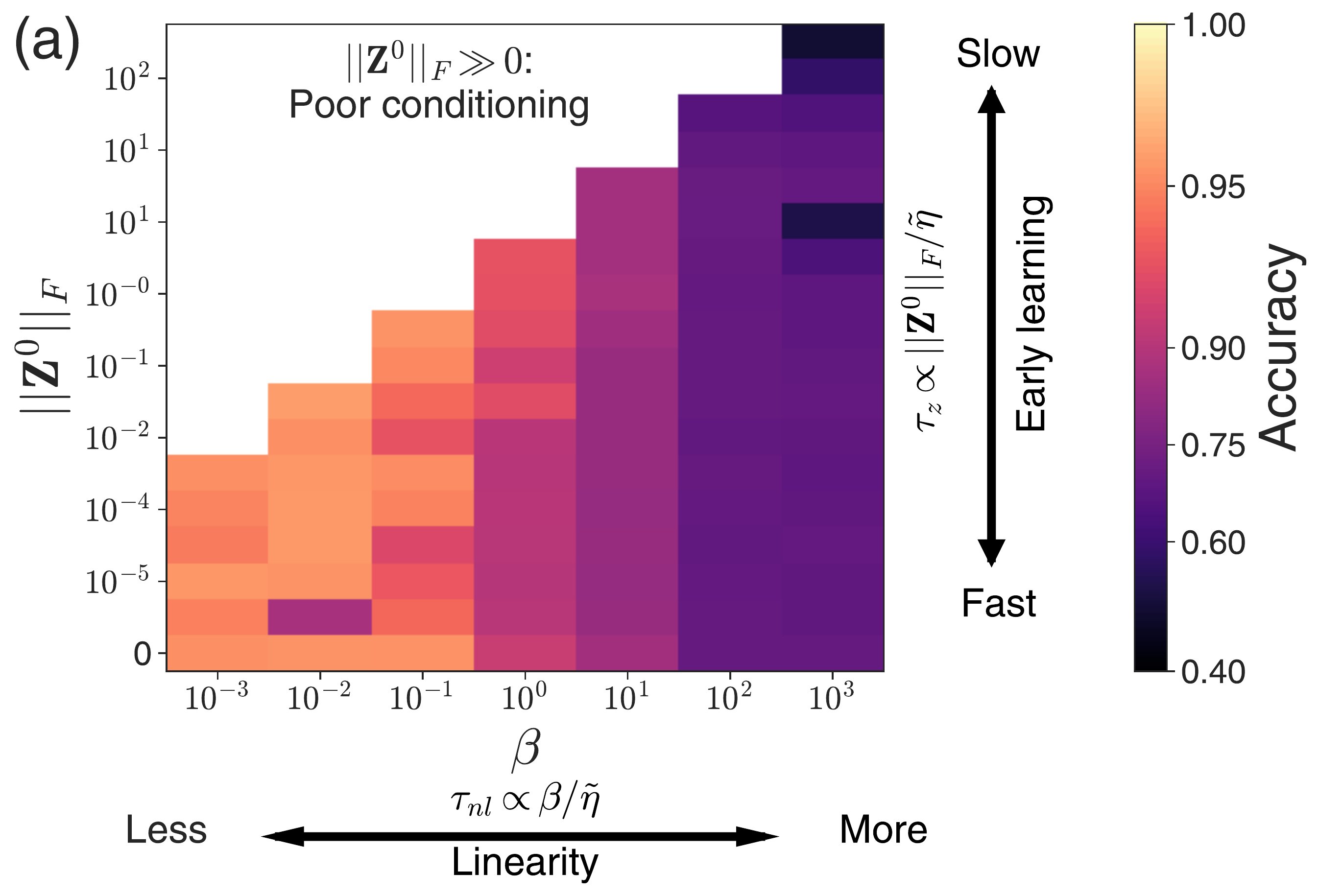}} &
\includegraphics[width=0.29\linewidth]{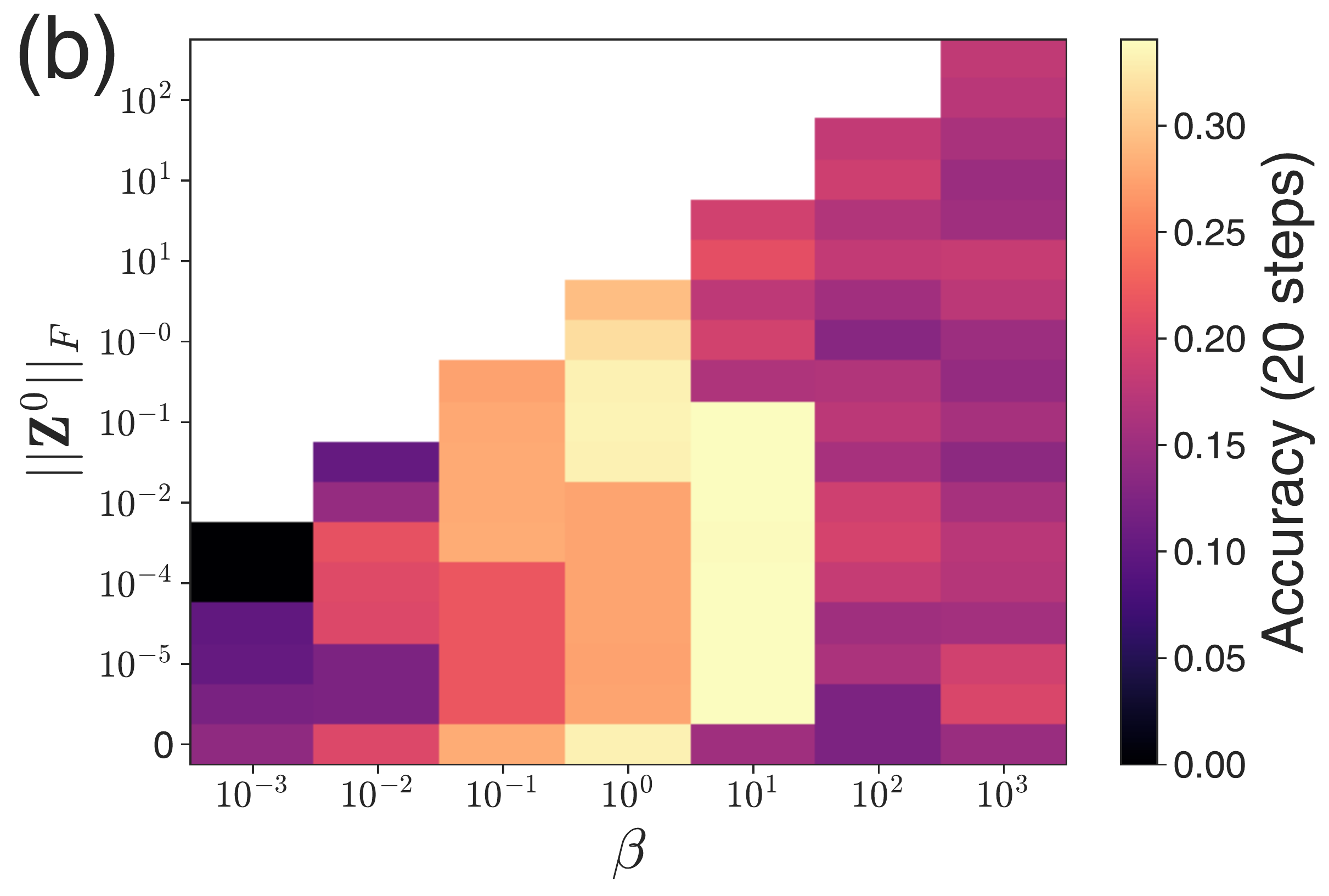}\\
& \includegraphics[width=0.29\linewidth]{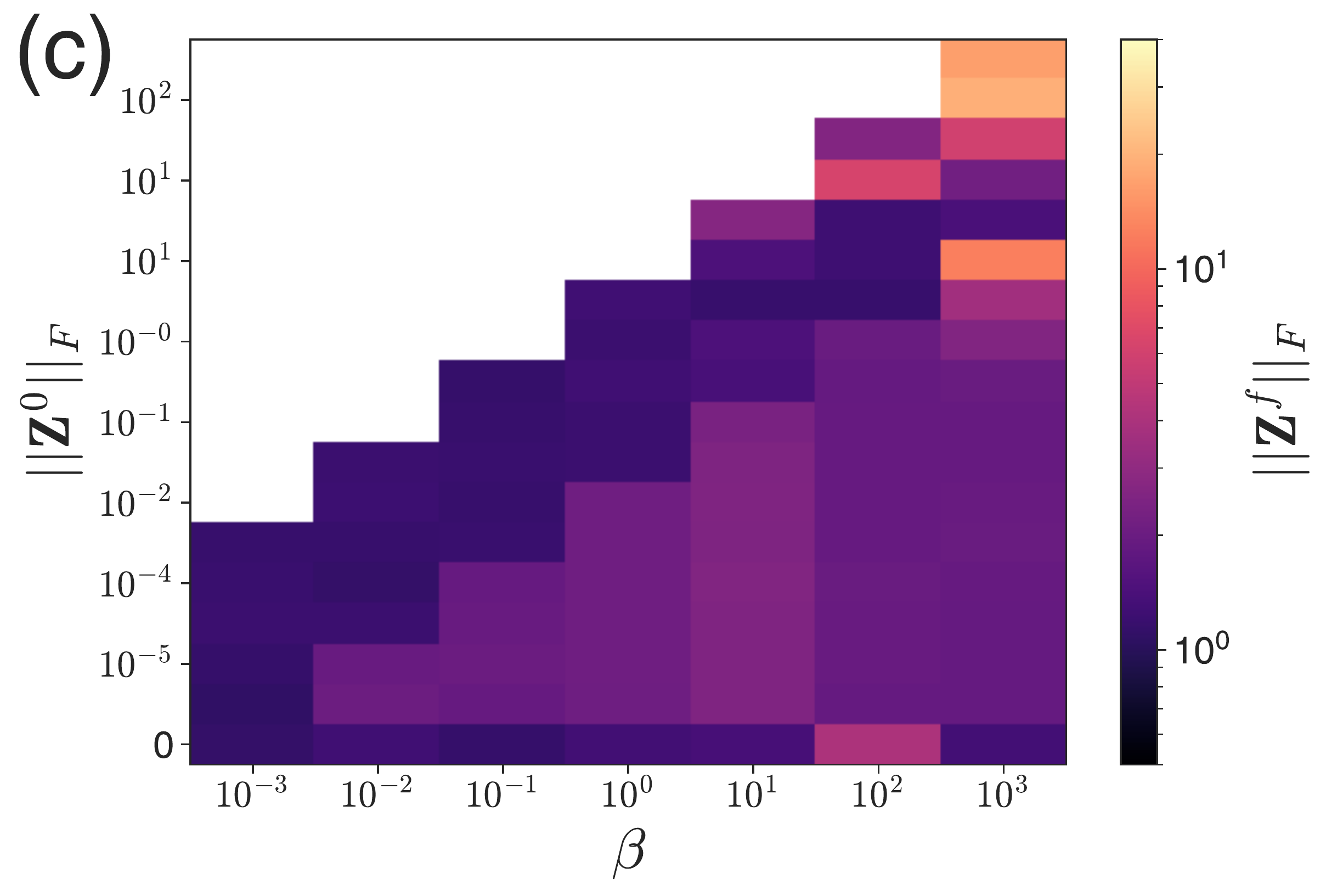}
\end{tabular}

\caption{Properties of early learning dynamics, which affect generalization,
can be determined by location in the $\beta$-$\lVert\Z^{0}\rVert_{F}$ phase plane (a).
At optimal learning rate $\lr^*$,
small $\beta$ and larger
$\lVert\Z^{0}\rVert_{F}$ leads to slower 
early learning (b), and larger $\beta$ increases time before nonlinear dynamics contributes
to learning.
Large $\lVert\Z^{0}\rVert_{F}$ has poorly conditioned linearized dynamics. Generalization
for a wide resnet trained on CIFAR10 is highly sensitive to $\beta$, and relatively
insensitive to $\lVert\Z^{0}\rVert_{F}$ outside poor conditioning regime. Final logit variance is relatively
insensitive to parameters (c).}
\label{fig:phase_plane_diagram}
\end{figure}

In Figure ~\ref{fig:phase_plane_diagram} (a), we show a schematic of the phase plane. The colormap shows the test performance of
a wide residual network \citep{zagoruyko_wide_2017}, without batch normalization, trained on CIFAR10 in different parts
of the phase plane. The value of $\beta$ makes a large difference in generalization, with optimal performance
achieved at $\beta\approx 10^{-2}$.
In general, larger $\beta$ performed worse than small $\beta$ as expected. Moreover, we observe similar generalization
for all sufficiently large $\beta$; this is to be expected since models in this regime are close to their 
linearization throughout training (see Figure \ref{fig:loss_collapse}) and we expect the linearized models to have 
$\beta$-independent performance. Generalization was largely insensitive to $\lVert\Z^{0}\rVert_{F}$ so long as the 
network was sufficiently well-conditioned to be trainable. This suggests that long term learning is insensitive to 
$\tz$.

In Figure~\ref{fig:phase_plane_diagram} (b), we plot the accuracy after 20 steps of optimization (with
the optimal learning rate). For fixed $\Zparam$, the training speed was slow for the smallest $\beta$ and then became 
faster with increasing $\beta$. For fixed $\beta$ the training speed was fastest for small $\Zparam$ and slowed as 
$\Zparam$ increased. Both these phenomena were predicted by our theory and shows that both parameters are
important in determining the early-time dynamics.
However, we note that the relative accuracy across the phase plane at
early times did not correlate with the performance at late times. 

This highlights that differences in generalization are a dynamical phenomenon.
Another indication of this fact is that at the end of training, at time $t_{f}$,
the final training set logit
values $\lVert\Z^{f}\rVert_{F} \equiv \lVert\Z(\X,t_{f})\rVert_{F}$ tend towards $1$
independent of the initial  $\beta$ and $\Zparam$ (Figure \ref{fig:phase_plane_diagram}, 
(c)).
With the exception of the poorly-performing large $\lVert\Z^{0}\rVert_{F}$ regime,
the different models reach similar levels of certainty by the end of training, despite
having different generalization performances.
Therefore generalization is not well correlated with the final model
certainty (a typical motivation for tuning $\beta$).

\subsection{Architecture Dependence of the Optimal $\beta$}

Having demonstrated that $\beta$ controls the generalization performance of neural networks with 
softmax-cross-entropy loss, we now discuss the question of choosing the optimal $\beta$. Here we investigate this 
question through the lens of a number of different architectures. We find the optimal 
choice of $\beta$ to be strongly architecture dependent. Whether or not the optimal $\beta$ can be predicted analytically is an open question that we leave for future work.
Nonetheless, we show that all architectures
considered display optimal $\beta$ between approximately $10^{-2}$ and $10^1$. We observe that by taking the
time to tune $\beta$ it is often the case that performance can be improved over the naive setting of $\beta = 1$.  

\subsubsection{Wide Resnet on CIFAR10}

\begin{figure}[h]
\centering
\includegraphics[width=0.32\linewidth]{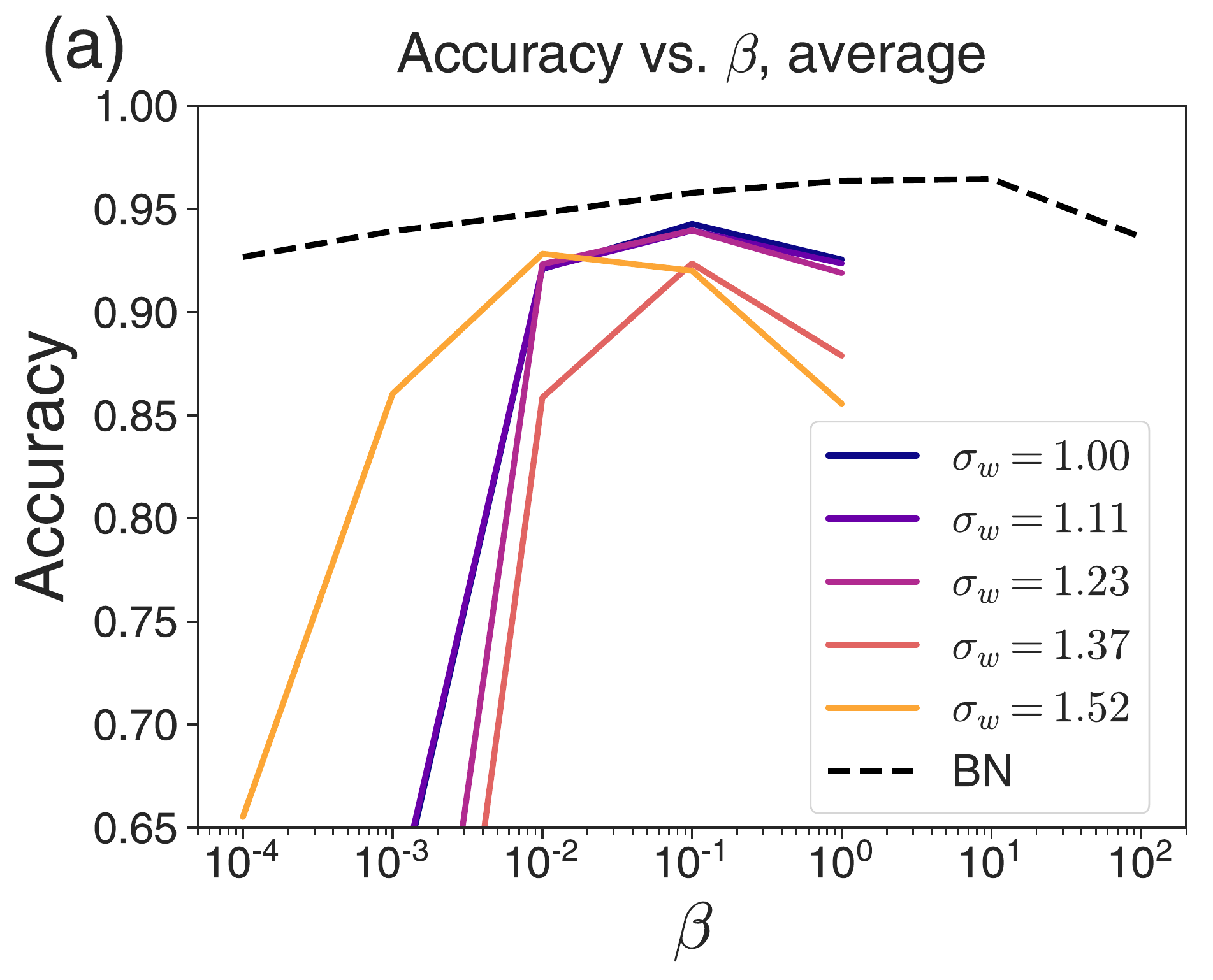}
\includegraphics[width=0.32\linewidth]{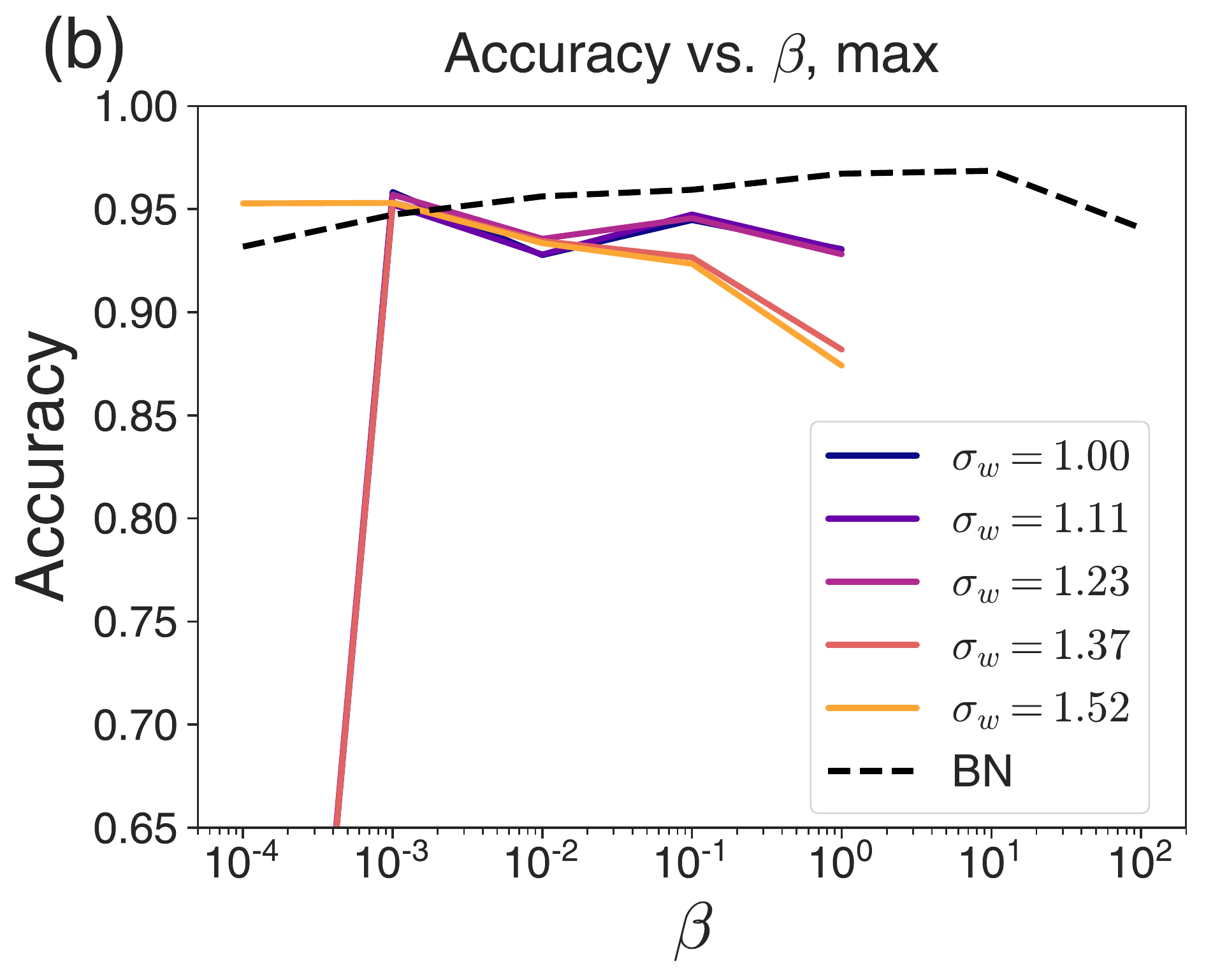}
\includegraphics[width=0.32\linewidth]{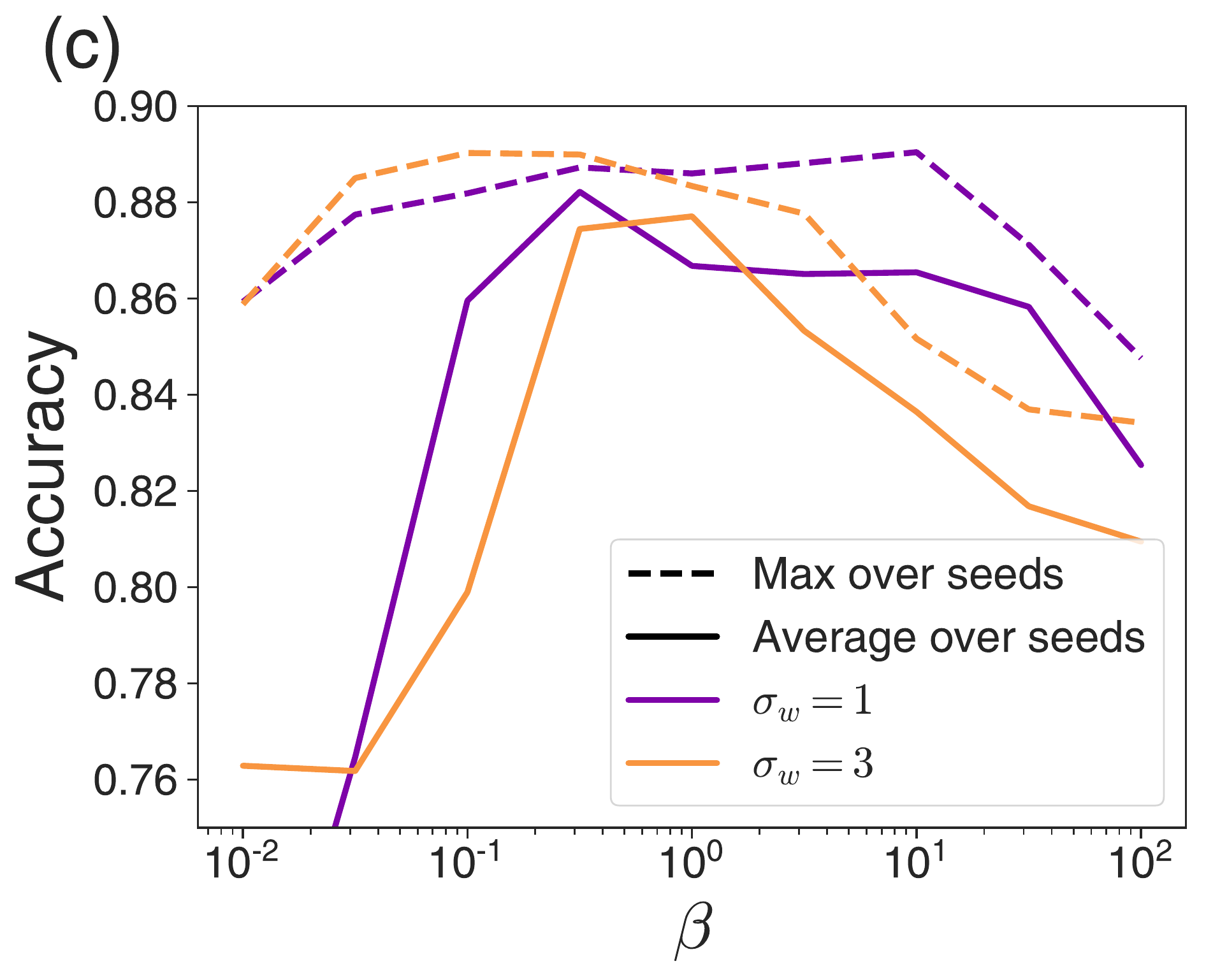}
\caption{Dependence of test accuracy for various architectures with $\beta$ tuning.
(a) For WRN with batchnorm, trained on CIFAR10, the optimal $\beta\approx 10$. Without batchnorm,
the performance of the network can be nearly recovered with $\beta$-scaling alone with $\beta\approx 10^{-2}$.
Even poorly conditioned networks (achieved by increasing weight scale $\sigma_w$) recover performance.
(b) For $\beta<10^{-2}$, learning is less stable, as evidenced by low average performance but high
maximum performance (over $10$ random seeds). (c) We see similar phenomenology on the IMDB sentiment analysis
task trained with GRUs - where average-case best performance is near $\beta = 1$ but peak performance
is at small $\beta$.}
\label{fig:bn_comparison}
\end{figure}

In Figure~\ref{fig:bn_comparison} (a)
we show the accuracy against $\beta$ for several wide residual networks whose 
weights are drawn from normal distributions of different variances, $\sigma_w^2$, trained without batchnorm,
as well as a network with $\sigma_w^{2} = 1$ trained with batchnorm (averaged over $10$ seeds).
The best average performance
is attained for $\beta<1$, $\sigma_w = 1$ without batchnorm, and in particular networks with large
$\sigma_w$ are dramatically improved with $\beta$ tuning. The network with batchnorm is better at all
$\beta$, with optimal $\beta\approx 10$. However, we see that the best performing seed is often at a lower
$\beta$ (Figure~\ref{fig:bn_comparison} (b)), with larger $\sigma_w$ networks competitive with $\sigma_w = 1$,
and even with batchnorm at fixed $\beta$ (though batchnorm with $\beta = 10$ still performs the best).
This suggests that small $\beta$ can improve best case performance, at the cost of stability.
Our results emphasize the importance of tuning $\beta$, especially for
models that have not otherwise been optimized.

\subsubsection{Resnet50 on ImageNet}

\begin{table}[h]
\caption{Accuracy on Imagenet dataset for ResNet-50. Tuning $\beta$ significantly improves accuracy.}
\begin{center}
\begin{tabular}{l c c} 
Method & Accuracy (\%)\\
 \hline
ResNet-50  \citep{ghiasi_dropblock_2018} & $76.51 \pm 0.07$\\
ResNet-50 + Dropout \citep{ghiasi_dropblock_2018} & $76.80 \pm 0.04$\\
ResNet-50 + Label Smoothing \citep{ghiasi_dropblock_2018} & $77.17 \pm 0.05$\\
ResNet-50 + Temperature check ($\beta=0.3$) & $77.37 \pm 0.02$\\
\end{tabular}
\end{center}
\label{tbl:resnet50}
\end{table}

Motivated by our results on CIFAR10, we experimentally explored the effects of $\beta$ as a tunable
hyperparameter for ResNet-50 trained on Imagenet.
We follow the experimental protocol established by \citep{ghiasi_dropblock_2018}. A key 
difference between this procedure and standard training is that we train for substantially longer: the number of training epochs is increased from $90$ to 
$270$. \citet{ghiasi_dropblock_2018} found that this longer training regimen was beneficial when using
additional regularization. Table \ref{tbl:resnet50} shows that scaling $\beta$ improves accuracy for ResNet-50 with batchnorm. 
However, we did not find that using $\beta<1$ was optimal for ResNet-50 without normalization. This further emphasizes the subtle architecture dependence that warrants further study.

\subsubsection{GRUs on IMDB Sentiment Analysis}

To further explore the architecture dependence of optimal $\beta$, we train GRUs
(from \citet{maheswaranathan_reverse_2019})
whose weights are drawn from 
two different distributions on an IMDB sentiment analysis task that has been widely 
studied~\citep{maas_learning_2011}. We plot the results in Figure~\ref{fig:bn_comparison} (c) and observe that the 
results look qualitatively similar to the results on CIFAR10 without batch normalization. We observe a peak 
performance near $\beta \sim 1$ averaged over an ensemble of networks, but we observe that smaller $\beta$ can give 
better optimal performance at the expense of stability.  

\section{Conclusions}

Our empirical results show that tuning $\beta$ can yield sometimes significant improvements to model performance. Perhaps most surprisingly, we observe gains on ImageNet even with the highly-optimized 
ResNet50 model.
Our results on CIFAR10 suggest that the effect of $\beta$ may be even stronger in networks which are 
not yet
highly-optimized, and results on IMDB show that this effect holds beyond the image classification setting.
It is possible that even more gains can be made by more carefully tuning $\beta$ jointly with other hyperparameters, in particular the
learning rate schedule and batch size.

One key lesson of our theoretical work is that properties of learning dynamics must be compared using the right units. 
For
example, $\tnl\propto 1/\beta\lr$, which at first glance suggests that models with smaller $\beta$ will become 
nonlinear
more slowly than their large $\beta$ counterparts. However, analyzing $\tnl$
with respect to the effective learning rate $\tlr = \beta^2\lr$ yields $\tnl\propto \beta/\tlr$. Thus we see that, in 
fact, networks with smaller $\beta$  tend to become more non-linearized before much learning has occurred,
compared to networks with large
$\beta$ which can remain in the linearized regime throughout training. Our numerical results confirm
this intuition developed using the theoretical analysis.


As discussed above, our analysis does not predict the optimal $\beta$ or $\lr$. Extending the theoretical results to make predictions about these quantities is an interesting avenue for future work. Another area that warrants further study is the instability in training at small $\beta$.

\clearpage




\bibliography{aa_refs_neurips2020}

\medskip

\clearpage

\appendix

\section{Linearized learning dynamics}

\label{app:linearized_dyn}

\subsection{Fixed points}

For the linearized learning dynamics, the trajectory $\z(\x,t)$ can be written in terms of the trajectories of the
training set as
\begin{equation}
\z(\x,t)-\z(\x,0) = \ntk(\x,\X)\ntk^{+}(\X,\X)(\z(\X,t)-\z(\X,0))
\end{equation}
where $+$ is the pseudo-inverse. Therefore, if one can solve for $\z(\X,t)$, then in principle properties
of generalization are computable.

However, in general Equation \ref{eq:z_ntk} does not admit an analytic solution even for fixed $\ntk$, in contrast to the case of
mean squared loss. It not even have an equilibrium - if the model can
achieve perfect training accuracy, the logits will grow indefinitely. However, there is a guaranteed
fixed point if the appropriate
$L_2$ regularization is added to the training objective. Given a regularizer $\frac{1}{2}\lambda_{\th}\lVert\delta\th\rVert^2$
on the change in parameters $\delta\th = \th(t)-\th(0)$,
the dynamics in the linearized regime are given by
\begin{equation}
\dot{\z}(\x) = \beta\lr\ntk(\x,\X)(\Y(\X)-\soft(\beta\z(\X)))-\lambda_{\th}\delta\z(\x)
\label{eq:ntk_with_reg}
\end{equation}
where the last term comes from the fact that $\frac{\partial \z}{\partial\th}\delta\th = \dz(\x)$ in the linearized limit.

We can write down self-consistent equations for equilibria, which are
approximately solvable in certain limits. For an arbitrary input $\x$, the equilibrium solution
$\zs(\x)$ is
\begin{equation}
0 = \beta\ntk(\x,\X)(\Y(\X)-\soft(\beta\zs(\X)))-\lambda_{\th}\delta\zs(\x)
\end{equation}
This can be rewritten in terms of the training set as
\begin{equation}
\delta\zs(\x) = \ntk(\x,\X)\ntk^{+}(\X,\X)\zs(\X)
\end{equation}
similar to kernel learning.

It remains then to solve for $\zs(\X)$. We have:
\begin{equation}
\delta\zs(\X) = \frac{\beta}{\lambda_{\th}}\ntk(\X,\X)[\Y(\X)-\soft(\beta\zs(\X))]
\end{equation}
We immediately note that the solution depends on the initialization. We assume $\z(\x,0) = 0$,
so $\delta\z = \z$ in order to simplify the analysis.
The easiest case to analyze is when $\lVert\beta\zs(\X)\rVert_{F}\ll 1$. Then we have:
\begin{equation}
\zs(\X) = \frac{\beta}{\lambda_{\th}}\ntk(\X,\X)\left[\Y(\X)-\frac{1}{\K}(1+\beta\zs(\X))\right]
\end{equation}
which gives us
\begin{equation}
\zs(\X) = \frac{\beta}{\lambda_{\th}}\left[1+\frac{\beta}{K\lambda_{\th}}\ntk(\X,\X)\right]^{-1}\ntk(\X,\X)(\Y(\X)-1/\K)
\end{equation}
Therefore the self-consistency condition for this solution is $\lVert\frac{\beta}{\lambda_{\th}}\ntk\rVert_{F}\ll 1$, which simplifies the solution
to
\begin{equation}
\zs(\X) = \frac{\beta}{\lambda_{\th}}\ntk(\X,\X)(\Y(\X)-1/\K)
\end{equation}
This is equivalent to the solution after a single step of (full-batch) SGD with appropriate learning rate.
We note that unlike linearized dynamics with $L_2$ loss and a full-rank kernel,
there is no guarantee that the solution converges to $0$ training error.

The other natural limit is $\lVert\beta\zs(\X)\rVert_{2}\gg1$. We focus on the $2$ class case, in order to take advantage of
the conserved quantity of learning with cross-entropy loss. We note that the vector on the right hand side
of Equation \ref{eq:z_ntk} sums to $1$ for every training point.  Suppose at initialization, $\ntk_{\th}$
has no logit-logit interactions, as is the case for most architectures in the infinite width limit with
random initialization. More formally, we can write $\ntk_{\th} = \m{Id}_{\K\times\K}\otimes \ntk_{x}$ where
$\ntk_{x}$ is $\M\times\M$. Then, the sum of the logits for any input
$\x$ is conserved during linearized training, as we have:
\begin{equation}
\m{1}^{\tpose}\dot{\z}(\x) = \lr\beta\m{1}^{\tpose}\left[\m{Id}_{\K\times\K}\otimes \ntk_{x}\right](\Y-\soft(\beta\z(\X)))
\end{equation}
Multiplying the right hand side through, we get
\begin{equation}
\m{1}^{\tpose}\dot{\z}(\x) = \lr\beta \ntk_{x}\left[\m{1}^{\tpose}(\Y-\soft(\beta\z(\X)))\right] = 0
\end{equation}
(Note that if $\ntk_{\th}$ has explicit dependence on the logits, there still is a conserved quantity, which
is more complicated to compute.)

Now we can analyze $\lVert\beta\zs(\X)\rVert_{F}\gg1$. With two classes, and $\z(\X) = 0$ at initialization,
we have $\zs_{1} = -\zs_{2}$. Therefore, without loss of generality, we
focus on $\zs_{1}$, the logit of the first class.
In this limit, the leading order correction to the softmax is approximately:
\begin{equation}
\soft(\beta\zs_1) \approx \m{1}_{\zs_{1}>0}-\sign(\zs_{1})e^{-2\beta|\zs_{1}|}
\end{equation}
The self-consistency equation is then:
\begin{equation}
\zs_1(\X) = \frac{\beta}{\lambda_{\th}}\ntk(\X,\X)\left[\Y(\X)-\m{1}_{\zs_{1}>0}+\sign(\zs_{1})e^{-2\beta|\zs_{1}|}\right]
\end{equation}
The vector on the right hand side has entries that are $O(e^{-2\beta|\zs_{1}|})$ for correct classifications, and $O(1)$
for incorrect ones. If we assume that the training error is $0$, then we have:
\begin{equation}
\zs_1(\X) = \frac{\beta}{\lambda_{\th}}\ntk(\X,\X)\sign(\zs_{1})e^{-2\beta|\zs_{1}|}
\label{eq:large_betaz}
\end{equation}
This is still non-trivial to solve, but we see that the self consistency condition is that $\ln(\beta||\ntk||_{F}/\lambda_{\th})\gg1$.

Here also it may be difficult to train and generalize well. The individual elements of the right-hand-side vector
are broadly distributed due to the exponential - so the outputs of the model are sensitive to/may only depend on
a small number of datapoints. Even if the equilibrium solution has no training loss, generalization error may be high for the same
reasons.

This suggests that even for NTK learning (with $L_2$ regularization), the scale of $||\beta\z||$ plays an important role
in both good training accuracy and good generalization. In the NTK regime, there is one unique solution so (in the continuous
time limit) the initialization doesn't matter; rather, the ratio of $\beta$ and $\lambda_{\th}$ (compared to the appropriate
norm of $\ntk$) needs to be balanced to prevent falling into the small $\beta\z$ regime (where training error might be large)
or the large $\beta\z$ regime (where a few datapoints might dominate and reduce generalization).

\subsection{Dynamics near equilibrium}

The dynamics near the equilibrium can be analyzed by expanding around the fixed point equation.
We focus on the dynamics on the training set. The dynamics of the difference $\tilde{\z}(\X) = \z(\X)-\z^*(\X)$
for small perturbations is given by
\begin{equation}
\dot{\tilde{\z}}(\X) = -\lr\left[\beta^2[\ntkz(\X,\X)]\dsoft(\beta\z^*(\X))+\lambda_{\th}\right]\tilde{\z}(\X)
\label{eq:linearized_eq}
\end{equation}
where $\dsoft$ is the derivative of the softmax matrix
\begin{equation}
\dsoft(\z)\equiv\frac{\partial \soft(\z)}{\partial\z'} = \m{diag}(\soft(\z))-\soft(\z)\soft(\z')^{\tpose}
\label{eq:logit_derivative}
\end{equation}

We can perform some analysis
in the large and small $\beta$ cases (once again ignoring $\lambda_{z}$).
For small $\lVert\beta\zs(\X)\rVert_{F}$, we have $\lVert\frac{\beta}{\lambda_{\th}}\ntk\rVert_{F}\ll 1$ which leads to:
\begin{equation}
\dsoft(\beta\z^*(\X)) = (1/\K-\m{1}\m{1}^{\tpose}/\K^2)+O(\beta\ntk)
\end{equation}
This matrix has $K-1$ eigenvalues with value $1/\K$, and one zero eigenvalue (corresponding to the conservation
of probability). Therefore $\lVert\beta^2[\ntkz(\X,\X)]\dsoft(\beta\z^*(\X))\rVert_{F}\ll \lambda_{\th}$, and the well-conditioned
regularizer dominates the approach to equilibrium.

In the large $\beta$ case ($\ln(\beta||\ntk||/\lambda_{\th})\gg1$),
the values of $\soft(\beta\z(\X))$ are exponentially close to $0$ ($K-1$ values) or $1$ (the value corresponding
to the largest logit).
This means that
$\dsoft(\beta\z(\X))$ has exponentially small values in $\lVert\beta\z(\X)\rVert_{F}$ - if any one of
$\soft(\beta\z_i(\X))$ and $\soft(\beta\z_j(\X))$
is exponentially small, the corresponding element of $\dsoft(\beta\z(\X))$ is as well; for the largest logit $i$
the diagonal is $\soft(\beta\z_i(\X))(1-\soft(\beta\z_i(\X)))$ which is also exponentially small.

From Equation \ref{eq:large_betaz}, we have $\lambda_{\th}\ll\beta^2e^{2\beta|\zs_{1}|}$; therefore, though
the $\dsoft$ term of $\H$ is exponentially small, it dominates the linearized dynamics near the fixed point,
and the approach to equilibrium is slow.
We will analyze the conditioning of the dynamics in the remainder of this section.

\subsection{Conditioning of dynamics}

Understanding the conditioning of the linearized dynamics
requires understanding the spectrum of the Hessian matrix
$\H = \left(\m{Id}_{\z}\otimes \ntk(\X,\X)\right) \dsoft(\beta\zs(\X))$. In the limit of large model size,
the first factor is
block-diagonal with training set by training set blocks (no logit-logit interactions), and the second term
is block-diagonal with $\K\times\K$ blocks (no datapoint-datapoint interactions).

We will use the following lemma to get bounds on the conditioning:

\textbf{Lemma:} Let $\m{M} = \m{A}\m{B}$ be a matrix that is the product of two matrices.
The condition number $\kappa(\m{M})\equiv \frac{\lambda_{\m{M},\max}}{\lambda_{\m{M},\min}}$ has bound
\begin{equation}
\kappa(\m{B})/\kappa(\m{A})\leq \kappa(\m{M})\leq \kappa(\m{A})\kappa(\m{B})
\label{eq:kappa_bound}
\end{equation}

\textbf{Proof:}
Consider the vector $\m{v}$ that is the eigenvector of $\m{B}$ associated
with $\lambda_{\m{B}, \min}$. Note that $||\m{A}\m{v}||/||\m{v}||\leq \lambda_{\m{A},\max}$.
Analogously, for $\m{w}$, the eigenvector associated with
$\lambda_{\m{B}, \max}$, $||\m{A}\m{w}||/||\m{w}||\geq \lambda_{\m{A},\min}$. This gives us the
two bounds:
\begin{equation}
\lambda_{\m{M},\min}\leq \lambda_{\m{A},\max}\lambda_{\m{B},\min}, ~ \lambda_{\m{M},\max}\geq \lambda_{\m{A},\min}\lambda_{\m{B},\max}
\end{equation}
This means that the condition number $\kappa(\m{H})\equiv \frac{\lambda_{\m{M},\max}}{\lambda_{\m{M},\min}}$ is
bounded by
\begin{equation}
\kappa(\m{M}) \geq \frac{\lambda_{\m{A},\max}\lambda_{\m{B},\min}}{\lambda_{\m{A},\min}\lambda_{\m{B},\max}} =\kappa(\m{B})/\kappa(\m{A})
\end{equation}
In total, we have the bound of Equation \ref{eq:kappa_bound},
where the upper bound is trivial to prove. $\square$

In particular, this means that a poorly conditioned $\dsoft(\beta\zs(\X))$ will lead to poor conditioning of the
linearized dynamics if the NTK $\ntk(\X,\X)$ is (relatively) well conditioned.
This bound will be important in establishing the poor conditioning of the linearized dynamics for the large logit regime
$||\beta\z||\gg 1$.

\subsubsection{Small logit conditioning}

\label{sec:small_log_con}

For $\lVert\beta\zs(\X)\rVert_{F}\ll 1$, the Hessian $\H$ is
\begin{equation}
\H = \frac{1}{\K}\left(1-\frac{1}{\K}\m{1}\m{1}^{\tpose}\right)\otimes \ntk(\X,\X)
\end{equation}
Since $\H$ is the Kroenecker product of two matrices, the condition numbers multiply, and we have
\begin{equation}
\kappa(\H) = \kappa(\ntk)
\end{equation}
which is well-conditioned so long as the NTK is.
Regardless, the well-conditioned regularization due to $\lambda_{\th}$ dominates the
approach to equilibrium.

\subsubsection{Large logit conditioning}

\label{sec:large_log_con}

Now consider $\lVert\beta\zs(\X)\rVert_{F}\gg 1$. Here we will show that the linearized dynamics is poorly conditioned,
and that $\kappa(\H)$ is exponentially large in $\beta$.

We first try to understand $\dsoft(\beta\zs(\x))$ for an
individual $\x\in\X$. To $0$th order (in an as-of-yet-undefined expansion), $\dsoft$ is zero -
at large temperature the softmax returns either $0$ or $1$, which
by Equation \ref{eq:logit_derivative} gives $0$ in all entries. The size of the corrections end up being exponentially
dependent on $\lVert\beta\zs\rVert_{F}$; the entries will have broad, log-normal type distributions with magnitudes
which scale as $\exp(-\beta|\zs_{1}|)$. There will be two scaling regimes one with a small number of labels
in the sense $\sqrt{\beta}\gg\sqrt{\ln(\K)}$, where the largest logit dominates the statistics,
and one where the number of labels is large
(and the central limit theorem applies to the partition function). In both cases, however, there is still
exponential dependence on $\beta$; we will focus on the first which is easier to analyze and more realistic
(e.g. for $10^{6}$ labels “large” $\beta$ is only $\sim 15$).

Let $z_{1}$ be the largest of $\K$ logits, $z_{2}$ the second largest, and so on. Then using Equation
 \ref{eq:logit_derivative} we have:
\begin{equation}
(\dsoft)_{i1} = -e^{-\beta(z_1-z_i)}
\end{equation}
for $i\neq 1$,
\begin{equation}
(\dsoft)_{ij} = \delta_{ij}e^{-\beta(z_{1}-z_{i})}-e^{-\beta(2z_1-z_i-z_j)}
\end{equation}
for $i\neq j$ and
\begin{equation}
(\dsoft)_{11} = e^{-\beta (z_{1}-z_{2})}
\end{equation}
The eigenvectors and eigenvalues can be approximately computed as:
\begin{equation}
(\m{v}_{1})_{1} = \frac{1}{\sqrt{2}},~(\m{v}_{1})_{2} = -\frac{1}{\sqrt{2}}, ~\lambda_1 =   2e^{-\beta(z_1-z_2)}
\end{equation}
\begin{equation}
(\m{v}_{2})_{1} = \frac{1}{\sqrt{2}},~(\m{v}_{2})_{2} = \frac{1}{\sqrt{2}}, ~\lambda_2 =   -\frac{1}{2}e^{-2\beta(z_1-z_2)}
\end{equation}
and for $i>2$,
\begin{equation}
(\m{v}_{i})_{i} = 1,~(\m{v}_{i})_{1} = e^{-\beta(z_{2}-z_{i})},~\lambda_{i} = e^{-\beta(z_{1}-z_{i})}
\end{equation}
with all non-explicit eigenvector components $0$. This expansion is valid
provided that $\beta/\K\gg 1$ (so that $e^{\beta(z_1-z_i)}\gg e^{\beta(z_{1}-z_{i+1})}$).

Therefore the spectrum of any individual
block $\dsoft(\beta\z(x))$ is exponentially small in $\beta$.
Using the bound in the Lemma, we have:
\begin{equation}
\kappa(\H)\geq e^{O(\beta|\zs_{1}|)}/\kappa(\ntk(\X,\X))
\end{equation}
This is a very loose bound, as it assumes that the largest eigendirections of $\dsoft$
are aligned with the smallest eigendirections of $\ntkz$, and vice versa. It is possible
$\kappa(\H)$ is closer in magnitude to the upper bound $e^{\beta(z_{2}-z_{K})}\kappa(\ntk(\X,\X))$.

Regardless, $\kappa(\H)$ is exponentially large in $\beta$ - meaning that the conditioning is exponentially
poor for large $\lVert\beta\zs\rVert_{F}$.

\section{SGD and momentum rescalings}

\label{app:mom_rescaling}

\subsection{Discrete equations}

Consider full-batch SGD training. The update equations for the parameters $\th$ are:
\begin{equation}
\th_{t+1} = \th_{t}-\lr \nabla_{\th}\Lf
\end{equation}
We will denote $\gd_{t}\equiv \nabla_{\th}\Lf$ for ease of notation.

Training with momentum, the equations of motion are given by:
\begin{equation}
\v_{t+1} = (1-\g) \v_{t}-\gd_{t}
\end{equation}
\begin{equation}
\th_{t+1} = \th_{t}+\lr\v_{t+1}
\end{equation}
where $\g\in[0,1]$.

One key point to consider later will be the relative magnitude $\dth$ of updates to the parameters. For SGD,
the magnitude of updates is $\lr||\gd||$. For momentum with slowly-varying gradients
the magnitude is $\lr||\gd||/\g$.

\subsection{Continuous time equations}

We can write down the continuous time version of the learning dynamics as follows. For SGD, for small learning rates
we have:
\begin{equation}
\frac{d\th}{dt} = -\eta \gd
\end{equation}
For the momentum equations we have
\begin{equation}
\frac{d\v}{dt} = -\g \v-\gd
\end{equation}
\begin{equation}
\frac{d\th}{dt} = \lr \v
\end{equation}

From these equations, we can see that in the continuous time limit, there are coordinate transformations
which can be used to cause sets of trajectories with different parameters to collapse to a single trajectory.
SGD is the simplest, where rescaling time to $\tau \equiv \eta t$ causes learning curves to be
identical for all learning rates.

For momentum, instead of a single universal learning curve,
there is a one-parameter family of curves
controlled by the ratio $\T_{mom} \equiv \lr/\g^2$. Consider rescaling
time to $\tau = a t$ and $\nub = b\v$, where $a$ and $b$ will be chosen to put the equations in a canonical form.
In our new coordinates, we have
\begin{equation}
\frac{d\nub}{d\tau} = -(\g/a) \nub-(b/a)\gd
\end{equation}
\begin{equation}
\frac{d\th}{d\tau} = \lr \nub/(ab)
\end{equation}
The canonical form we choose is
\begin{equation}
\frac{d\nub}{d\tau} = -\lam \nub-\gd
\label{eq:can_mom}
\end{equation}
\begin{equation}
\frac{d\th}{d\tau} = \nub
\label{eq:can_param}
\end{equation}
From which we arrive at $a = b = \sqrt{\lr}$, which gives us $\lam = \g/\sqrt{\lr}$.

Note that this is not a unique canonical form; for example, if we fix a coefficient of $-1$ on $\nub$,
we end up with
\begin{equation}
\frac{d\nub}{d\tau} = - \nub-(\lr/\g^2)\gd
\end{equation}
\begin{equation}
\frac{d\th}{d\tau} = \nub
\end{equation}
with $a = \g$. This is a different time rescaling, but still controlled by $\T_{mom}$.

Working in the canonical form of Equations \ref{eq:can_mom} and \ref{eq:can_param}, we can analyze the dynamics.
One immediate question is the difference between $\lam\ll 1$ and $\lam\gg1$. We note that the integral equation
\begin{equation}
\nub(\tau) = \nub(0)+\int_{0}^{\tau} e^{-\lam(\tau-\tau')}\gd(\tau')d\tau'
\end{equation}
solves the differential equation for $\nub$. Therefore, for $\lam\gg1$, $\nub(t)$ only depends on the current value
$\gd(t)$ and we have $\nub(\tau)\approx \gd(\tau)/\lam$. Therefore, we have, approximately:
\begin{equation}
\frac{d\th}{d\tau} \approx \frac{1}{\lam}\gd 
\label{eq:fast_mom_approx}
\end{equation}
This means that for large $\lam$ all the curves will approximately collapse, with timescale given by
$\sqrt{\eta}\lam^{-1} = \g\lr$ (dynamics similar to SGD).

For $\lam\ll 1$, the momentum is essentially the integrated gradient across all time. If $\nub(0) = 0$,
then we have
\begin{equation}
\frac{d\th}{d\tau} \approx\int_{0}^{\tau} \gd(\tau')d\tau'
\end{equation}
In this limit, $\th(\tau)$ is the double integral of the gradient with respect to time.

Given the form of the
controlling parameter $\T_{mom}$, we can choose to parameterize $\g = \tilde{\g}\sqrt{\eta}$. Under this parameterization,
we have $\T_{mom} = \tilde{\g}^{2}$. The dynamical equations then become:
\begin{equation}
\frac{d\nub}{d\tau} = -\tilde{\g}\nub-\gd
\label{eq:dtvdt_rescale}
\end{equation}
\begin{equation}
\frac{d\th}{d\tau} = \nub
\label{eq:dtthdt_rescale}
\end{equation}
which automatically removes explicit dependence on $\lr$.

One particular regime of interest is the early-time dynamics, starting from $\nub(0) = 0$. Integrating directly, we have:
\begin{equation}
\th(\tau) = -\frac{1}{2}\gd\tau^2+\frac{1}{6}\tilde{\g}\gd\tau^{3} +\ldots
\end{equation}
This means that $\tau$ alone is the correct timescale for early learning, at least until $\tau\tilde{\g}\sim 1$ - 
which in the original parameters corresponds to $t \sim 1/\g$ (the time it takes for the momentum to be first
``fully integrated'').

\subsection{Detailed analysis of momentum timescales}

One important subtlety is that $1$ is not the correct value to compare $\lam$ to. The real timescale involved is the one
over which $\gd$ changes significantly. We can approximate this in the following way. Suppose that there is some relative
change $\frac{\dth}{||\th||}\sim c$ of the parameters that leads to an appreciable relative change in $\gd$. Then the timescale
over which $\th$ changes by that amount is the one we must compare $\lam$ to.

We can compute that timescale in the following way. We assume $\gd$ fixed for what follows. Therefore, Equation
\ref{eq:fast_mom_approx} approximately holds. The timescale $\tau_{c}$ of the change is then given by:
\begin{equation}
\frac{\dth}{||\th||} = \frac{1}{\lam}\frac{||\gd||}{||\th||}\tau_{c} \sim c
\end{equation}
which gives
\begin{equation}
\tau_{c} \sim c\lam||\th||/||\gd||
\end{equation}
In particular, this means that the approximation is good when $\lam\tau_{c}\gg1$, which gives
$\g^2/\lr \gg \frac{||\gd||}{||\th||}$ - the former being a function of the dynamical parameters, the latter
being a function of the geometry of $\Lf$ with respect to $\th$.

One consequence of this analysis is that if the $||\th||$ remains roughly constant, for fixed $\lr$ and $\g$,
late in learning when the gradients become small the dynamics shifts into the regime where $\lam$
is large, and we effectively have SGD.

\subsection{Connecting discrete and continuous time}

One use for the form of the continuous time rescalings is to use them to compare learning curves for the actual
discrete optimization that is performed with
different learning rates. For small learning rates, the curves are expected to collapse, while for larger learning
rates the deviations from the continuous expectation can be informative.

With momentum, we only have perfect collapse when $\g$ and $\lr$ are scaled together.
However, one
typical use case for momentum is to fix the parameter $\g$, and sweep through different learning rates.
With this setup, if $\gd$ is changing slowly compared to $\g$
(more precisely, $\g^2/\lr\gg ||\gd||/||\th||$), as may be the case at later training times,
the change in parameters from a single
step is $\dth\sim (\lr/\g))||\gd||$ and the rescaling of taking $t$ to $\lr t$ (as for SGD) collapses the dynamics.
Therefore given a collection of varying $\lr$, but fixed $\g$ curves, it is possible to get intermediate and late
time dynamics on the same scale.

However, at early times, while the momentum is still getting ``up to speed" (i.e. in the first $1/\g$ steps),
the appropriate timescale is $\lr^{-1/2}$. Therefore, in order to
get learning curves to collapse across different $\lr$ at early times, we need to 
rescale $\g$ with $\lr$ as implied by Equations \ref{eq:can_mom} and \ref{eq:can_param}.
Namely, one must fix $\tilde{\g}$ and rescale $\g = \tilde{\g}\sqrt{\lr}$. We note that,
since $\g <1$, this gives us a restriction $\lr<\tilde{\g}^{-2}$ for the maximum learning that can be supported
by the rescaled momentum parameter.

\subsection{Momentum equations with softmax-cross-entropy learning}

\label{app:param_change}

For cross-entropy learning with softmax inputs $\beta \z$, all the scales acquire dependence on $\beta$. If we define
$\ell_{z}\equiv \left|\left|\frac{\partial \Lf}{\partial \beta \z}\right|\right|$
and $g_{z}\equiv \left|\left|\frac{\partial \z}{\partial \th}\right|\right|_{F}$, then we have,
approximately, $||\gd|| \approx \beta \ell_{z}g_{z}$.

Consider the goal of obtaining identical early-time learning curves for different values of $\beta$.
(The curves are only globally consistent across $\beta$ in the linearized regime.)
In order to get learning curves to collapse,
we want $\frac{d\Lf}{d\tau}$ to be independent of $\beta$ in the rescaled time units. We note that the change
in the loss function $\dlf$ from a single step of SGD goes as
\begin{equation}
\dlf \sim \lr\beta^2\ell_{z}^{2}g_{z}^{2}
\end{equation}
This suggests that one way to collapse learning curves is to plot them against
the rescaled learning rate
$\tlr t$, where $\tlr = \lr\beta^2$.
While hyperparameter tuning across $\beta$, one could use $\lr = \tlr/\beta^{2}$,
sweeping over $\tlr$ in order to easily obtain comparable learning curves.

However, a better goal for a learning rate rescaling is to try and stay within the
continuous time limit - that is, to control the change in
parameters $\dth$ for a single step to be 
small across $\beta$. We have
\begin{equation}
\dth\sim \lr \beta\ell_{z}g_{z}
\end{equation}
which suggests that maximum allowable learning rates will scale as $1/\beta$.
This suggests setting $\lr = \hat{\lr}\beta^{-1}$, and rescaling time as $\hat{\lr}\beta$
in order to best explore the continuous learning dynamics.

We can perform a similar analysis for the momentum optimizers.
We begin by
analyzing the continuous time equations for the 
dynamics of the loss. Starting with the rescalings from Equations \ref{eq:dtvdt_rescale} and
\ref{eq:dtthdt_rescale} we have
\begin{equation}
\frac{d\nub}{d\tau} = -\tilde{\g}\nub-\beta\gd
\end{equation}
\begin{equation}
\frac{d\th}{d\tau} = \nub
\end{equation}
\begin{equation}
\frac{d\Lf}{d\tau} = \beta\gd\cdot \nub
\end{equation}
where $\gd = \frac{\partial \Lf}{\partial\beta\z}\frac{\partial\z}{\partial\th}$. Rescaling $\tau$
by $\beta$ gets us:
\begin{equation}
\frac{d\nub}{d\beta\tau} = -\frac{\tilde{\g}}{\beta}\nub-\gd
\end{equation}
\begin{equation}
\frac{d\th}{d\beta\tau} = \frac{1}{\beta}\nub
\end{equation}
\begin{equation}
\frac{d\Lf}{d\beta\tau} =\gd\cdot \nub
\end{equation}
This rescaling causes a collapse of the trajectories of the $\Lf$ at early times if $\tilde{\g}/\beta$
is constant for varying $\beta$.

One scheme to arrive at the final canonical form, across
$\beta$, is by the following definitions of $\lr$,
$\g$, and $\tau$:
\begin{itemize}
\item $\lr = \tlr \beta^{-2}$
\item $\g\equiv \beta\sqrt{\lr}\tilde{\g} = \sqrt{\tlr}\tilde{\g}$
\item $\tau\equiv \beta\sqrt{\lr}t = \sqrt{\tlr} t$
\end{itemize}
where curves with fixed $\tilde{\g}$ will collapse.
The latter two equations are similar to before, except with $\lr$ replaced with $\tlr$. The dynamical equations are
then:
\begin{equation}
\frac{d\nub}{d\tau} = -\tilde{\g}\nub-\gd
\end{equation}
\begin{equation}
\frac{d\th}{d\tau} = \frac{1}{\beta}\nub
\end{equation}
\begin{equation}
\frac{d\Lf}{d\tau} =\gd\cdot \nub
\end{equation}
The change in parameters from a single step (assuming constant $\gd$ and saturation) is
\begin{equation}
\dth  = \frac{||\gd||}{\tilde{\g}\beta}\sqrt{\hat{\lr}}
\end{equation}

If we instead want the change in parameters from a single step to be invariant of $\beta$
so the continuous time approximation holds, while maintaining collapse of trajectories, we
first note that
\begin{equation}
\dth\sim \frac{\sqrt{\lr}}{\tilde{\g}}\beta\ell_{z}g_{z}
\label{eq:dth_mom}
\end{equation}
from a single step of the momentum optimizer. To keep $\dth$ invariant of $\beta$,
we can set:
\begin{itemize}
\item $\lr = \hat{\lr} \beta^{-1}$
\item $\g\equiv \beta\sqrt{\lr}\tilde{\g} = \sqrt{\hat{\lr}}\tilde{\g} = \beta^{1/2}\sqrt{\hat{\lr}}\tilde{\g}$
\item $\tau\equiv \beta \sqrt{\lr}t = \beta^{1/2}\sqrt{\hat{\lr}} t$
\end{itemize}

Note that the relationship between $\g$ and $\tilde{\g}$ is the same in both schemes when
measured with respect to the raw learning rate $\lr$.

\section{Softmax-cross-entropy gradient magnitude}

\label{app:Z_dep}

\subsection{Magnitude of gradients in fully-connected networks}

The value of $\tz$ has nontrivial (but bounded) dependence on
$\lVert\Z^{0}\rVert_{F}$ via the $\lVert\ntk(\Y-\soft(\Z^{0}(\X)))\rVert_{F}$ term in Equation \ref{eq:tau_z_eq}. We can
confirm the dependence for highly overparameterized models by using the theoretical
$\ntk$.
In particular, for wide neural networks, the tangent kernel is block-diagonal
in the logits, and easily computable.

\begin{figure}[h]
\centering
\begin{tabular}{ccc}
\includegraphics[width=0.3\linewidth]{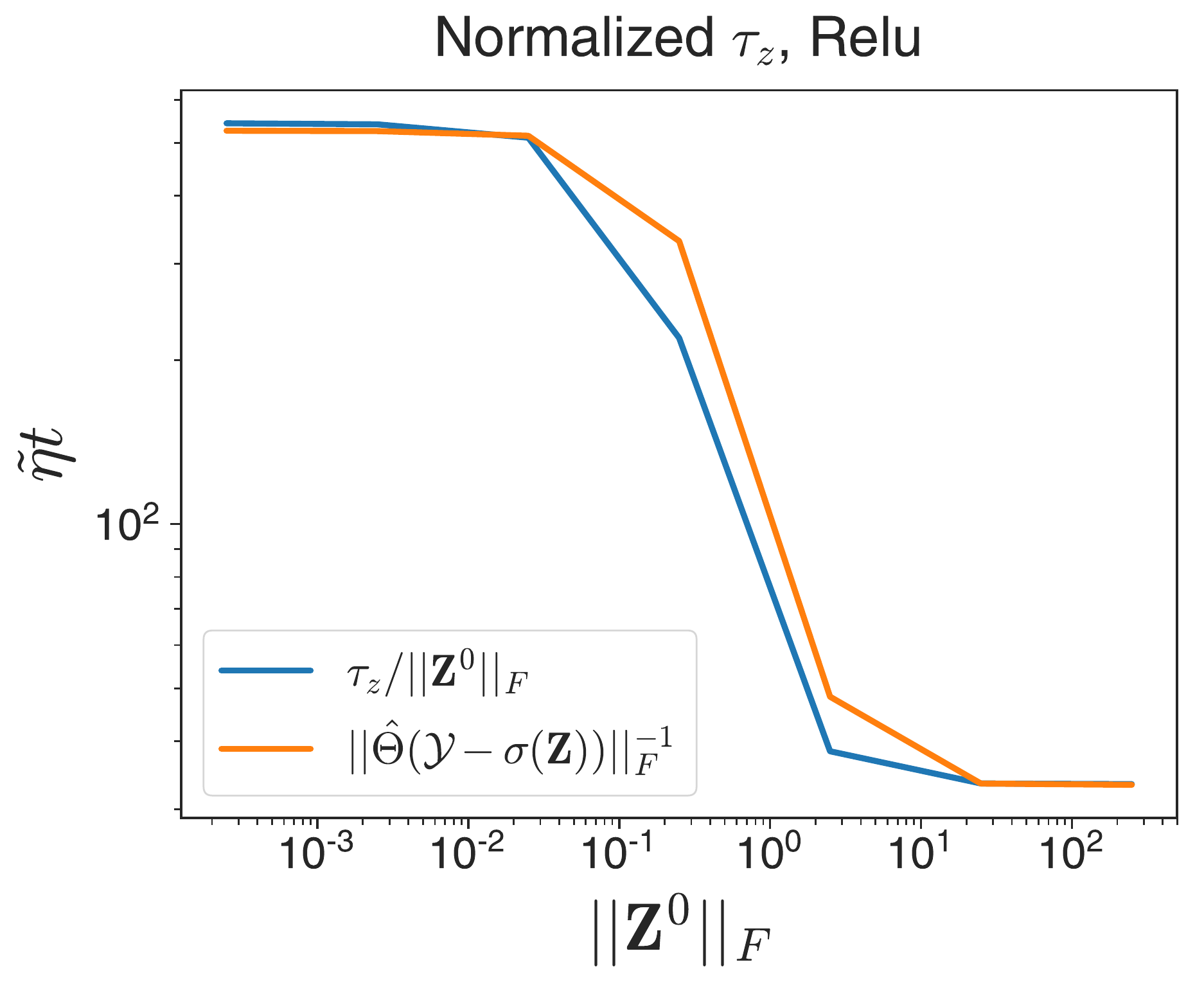} & 
\includegraphics[width=0.3\linewidth]{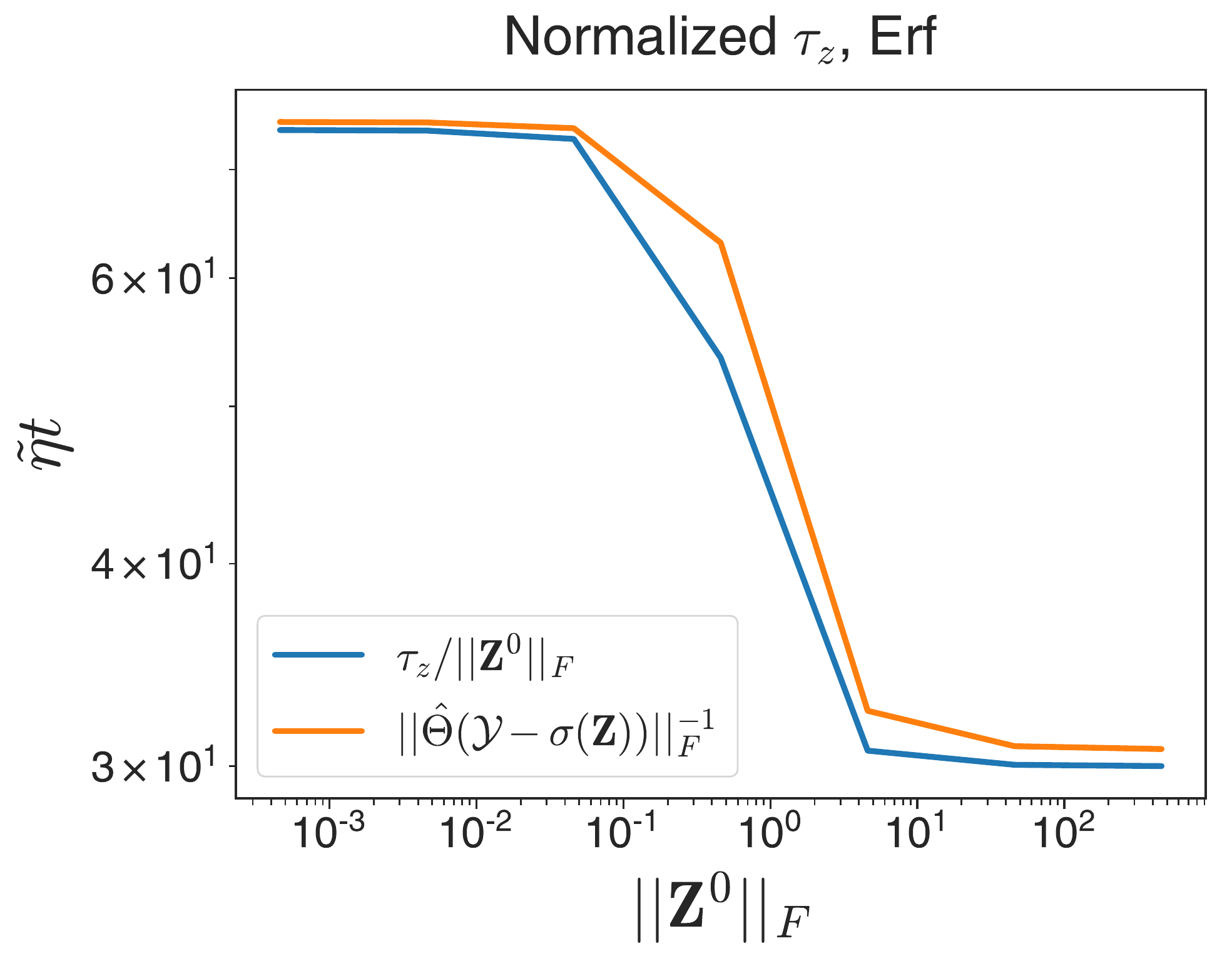} &
\includegraphics[width=0.3\linewidth]{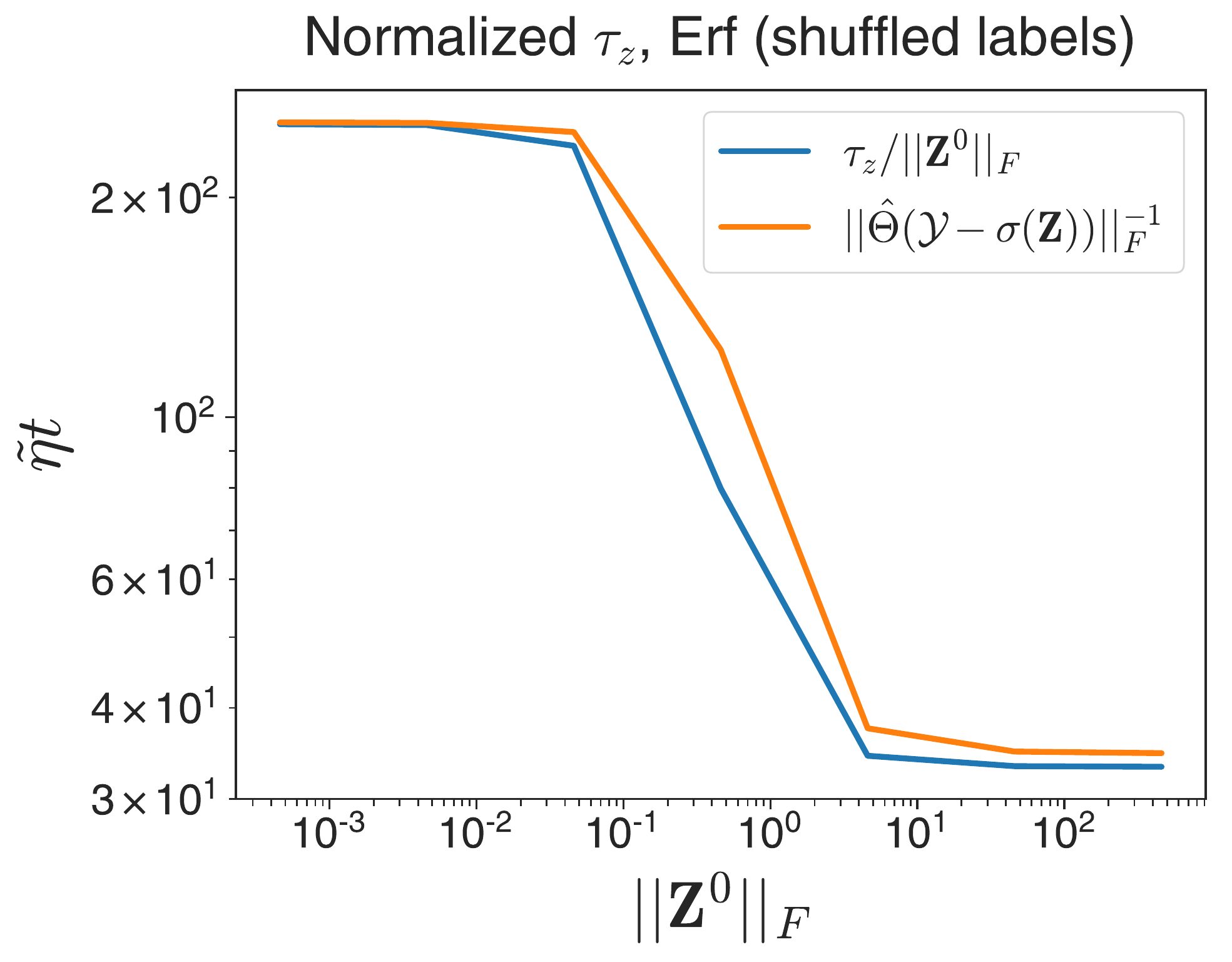}
\end{tabular}
\caption{$\tz/\Zparam$ is highly correlated with $\lVert\ntk(\Y-\soft(\Z^{0}))\rVert^{-1}$, with
$\ntk$ computed
in the infinite width limit
(in units of effective learning rate $\tlr = \beta^2\lr$). Ratio between normalized timescales at
large and small $\Zparam$ depends on nonlinearity (left and middle), as well as training
set statistics (right, CIFAR10 with shuffled labels).}
\label{fig:norm_tz}
\end{figure}

The numerically computed $\tz/\lVert\Z^{0}\rVert_{F}$ correlates well with $\lVert\ntk(\Y-\soft(\Z^{0}(\X)))\rVert_{F}^{-1}$
for wide ($2000$ hidden units/layer) fully connected networks (Figure \ref{fig:norm_tz}). 
The ratio depends on details like the nonlinearities in the network; for example,
Relu units tend to have a larger ratio than Erf units (left and middle).
The ratio also depends on the properties of the dataset. For example, the ratio
increases on CIFAR10 when the training labels are randomly shuffled (right).

Therefore in general the ratio of $\tz/\Zparam$ at large and small $\Zparam$ depends subtly
on the relationship between the NTK and the properties of the data distribution. A full
analysis of this relationship is beyond the scope of this work. The exact form of the
transition is likely even more sensitive to these properties and is therefore
harder to analyze than the ratio alone.

\section{Experimental details}

\subsection{Correlated initialization}

\label{app:cor_init}

In order to avoid confounding effects of changing
$\beta$ and $\lVert\Z^{0}\rVert_{F}$ with changes to $\ntk$, 
we use a
\emph{correlated initialization}
strategy (similar to \citep{chizat_lazy_2019}) which fixes $\ntk$ while allowing for independent
variation of $\beta$ and $\lVert\Z^{0}\rVert_{F}$.
Given a network with final hidden
layer $\h(\x,\th)$ and output weights $\m{W}_{O}$, we define a combined
network $\z_{\c}(\x,\tilde{\th})$ explicitly as
\begin{equation}
\z_{\c}(\x,\m{W}_{O,1},\m{W}_{O,2},\th_{1},\th_{2}) = \m{W}_{O,1}\h(\x,\th_{1}) +\m{W}_{O,2}\h(\x,\th_{2})
\end{equation}
where, at initialization, $\th_{1}=\th_{2}$, and the elements of
$\m{W}_{O,a}$ have statistics
\begin{equation}
\expect[(\W_{O,a})_{ij}(\W_{O,b})_{kl}] = \begin{cases}
\delta_{ik}\delta_{jl} &\text{~for~}a=b\\
\c\cdot\delta_{ik}\delta_{jl} &\text{~for~}a\neq b\\
\end{cases}
\end{equation}
for correlation coefficient $\c\in[-1,1]$, where $\delta_{ij}$ is the Kronecker-delta which is $1$ is $i = j$ and $0$ otherwise.
Under this approach, the initial magnitude of the training set logits is given by $\Zparam = \beta\sqrt{(1+\c)}\lVert\z^{0}\rVert_{F}$,
where $\lVert\z^{0}\rVert_{F}$ is the initial magnitude of the logits of the base model.
By manipulating $\beta$ and $\c$, we can
independently change $\beta$ and $\lVert\Z^{0}\rVert_{F}$ with the caveat that $\lVert\Z^{0}\rVert_{F}\leq\sqrt{2}\beta\lVert\z^{0}\rVert_{F}$ since
$\c\leq 1$. It follows that the small $\beta$, large $\lVert\Z^{0}\rVert_{F}$ region of the phase 
plane (upper left in Figure \ref{fig:phase_plane_diagram}) is inaccessible
with most well-conditioned models where $\lVert\z^{0}\rVert_{F}\sim 1$ at initialization.
If we only train one set of weights, the $\ntk$ is independent of $\c$.

Unless otherwise noted, all empirical studies in Sections \ref{sec:theory} and \ref{sec:late_time} involve training
a wide resnet on CIFAR10 with SGD, using GPUs, using the above correlated
initialization strategy to fix $\ntk$.
All experiments used JAX~\citep{bradbury_jax_2018} and experiments involving 
linearization or direct computation of the NTK used Neural 
Tangents~\citep{novak_neural_2019}. 

\end{document}